# An improved educational competition optimizer with multi-covariance learning operators for global optimization problems

Baoqi Zhao [1,3], Xiong Yang [2,*], Hoileong Lee [4] and Bowen Dong [5]

[1] Institute of Artificial Intelligence Application, Ningbo Polytechnic, Ningbo 315800, China;
[2] Zhicheng College, Fuzhou University, Fuzhou, 350002, China;
[3] School of Artificial Intelligence, Ningbo Polytechnic, Ningbo 315800, China;
[4] Faculty of Electronic Engineering & Technology, University Malaysia Perlis, 02600 Arau, Perlis, Malaysia;
[5] School of Electrical Automation and Information Engineering, Tianjin University, Tianjin 300072, China;
*Correspondence: 02116828@fdzcxy.edu.cn

**Abstract:** The educational competition optimizer is a recently introduced metaheuristic algorithm inspired by human behavior, originating from the dynamics of educational competition within society. Nonetheless, ECO faces constraints due to an imbalance between exploitation and exploration, rendering it susceptible to local optima and demonstrating restricted effectiveness in addressing complex optimization problems. To address these limitations, this study presents an enhanced educational competition optimizer (IECO-MCO) utilizing multi-covariance learning operators. In IECO, three distinct covariance learning operators are introduced to improve the performance of ECO. Each operator effectively balances exploitation and exploration while preventing premature convergence of the population. The effectiveness of IECO is assessed through benchmark functions derived from the CEC 2017 and CEC 2022 test suites, and its performance is compared with various basic and improved algorithms across different categories. The results demonstrate that IECO-MCO surpasses the basic ECO and other competing algorithms in convergence speed, stability, and the capability to avoid local optima. Furthermore, statistical analyses, including the Friedman test, Kruskal-Wallis test, and Wilcoxon rank-sum test, are conducted to validate the superiority of IECO-MCO over the compared algorithms. Compared with the basic algorithm (improved algorithm), IECO-MCO achieved an average ranking of 2.213 (2.488) on the CE2017 and CEC2022 test suites. Additionally, the practical applicability of the proposed IECO-MCO algorithm is verified by solving constrained optimization problems. The experimental outcomes demonstrate the superior performance of IECO-MCO in tackling intricate optimization problems, underscoring its robustness and practical effectiveness in real-world scenarios.

## 1. Introduction

The swift progression of artificial intelligence has become a cornerstone of societal advancement, complementing the ongoing development of various sectors. As the world evolves, it generates increasingly intricate demands across disciplines such as natural sciences, medicine, engineering, and economics, introducing challenges of unparalleled complexity [1]. Concurrently, continuous innovation in science and technology has given rise to a diverse array of AI-based methodologies, offering robust tools to address these multifaceted issues. With its exceptional capabilities in data analysis, adaptive learning, and logical reasoning, AI demonstrates significant potential for application across numerous fields [2, 3]. Optimization problems represent a critical area of study in scientific and engineering research, aiming to identify optimal solutions within constrained environments or limited resources. These problems often exhibit nonlinear dynamics, nonconvex structures, and high-dimensional spaces, accompanied by inherent complexity and uncertainty. Traditional deterministic methods and metaheuristic strategies are two predominant approaches employed to tackle such challenges [4]. However, deterministic techniques are frequently hindered by their computational inefficiency, particularly when addressing problems with extensive constraints and variables. Moreover, they are prone to converging to local optima, a significant limitation in real-world engineering applications [5]. The exponential growth in computational power and the continuous refinement of optimization techniques have catalyzed the emergence of metaheuristic algorithms. These stochastic methods are celebrated for their simplicity, robust ability to escape local optima, and independence from gradient-based information, making them highly effective for solving complex optimization problems [6]. Metaheuristic optimization techniques have gained popularity as a versatile class of methods, with widespread applications across diverse industries and fields such as image segmentation [7–9], task planning [10–12], structural optimization [13–15], energy system optimization [16–18], feature selection [19–21] and portfolio optimization [22, 23].

Over time, metaheuristic algorithms have diversified into several distinct branches. Researchers such as Xie and Xue have classified them into three main categories: evolution-based algorithms, physics-based algorithms, and swarm-based algorithms [24, 25]. Additionally, scholars like Jia and Abualigah have introduced a fourth category: algorithms modeled on human behavior [21, 22]. In this work, metaheuristic algorithms are categorized into five main classes (see Figure 1): evolutionary-based, swarm-based, physics-based, mathematically-based, and human-based approaches.

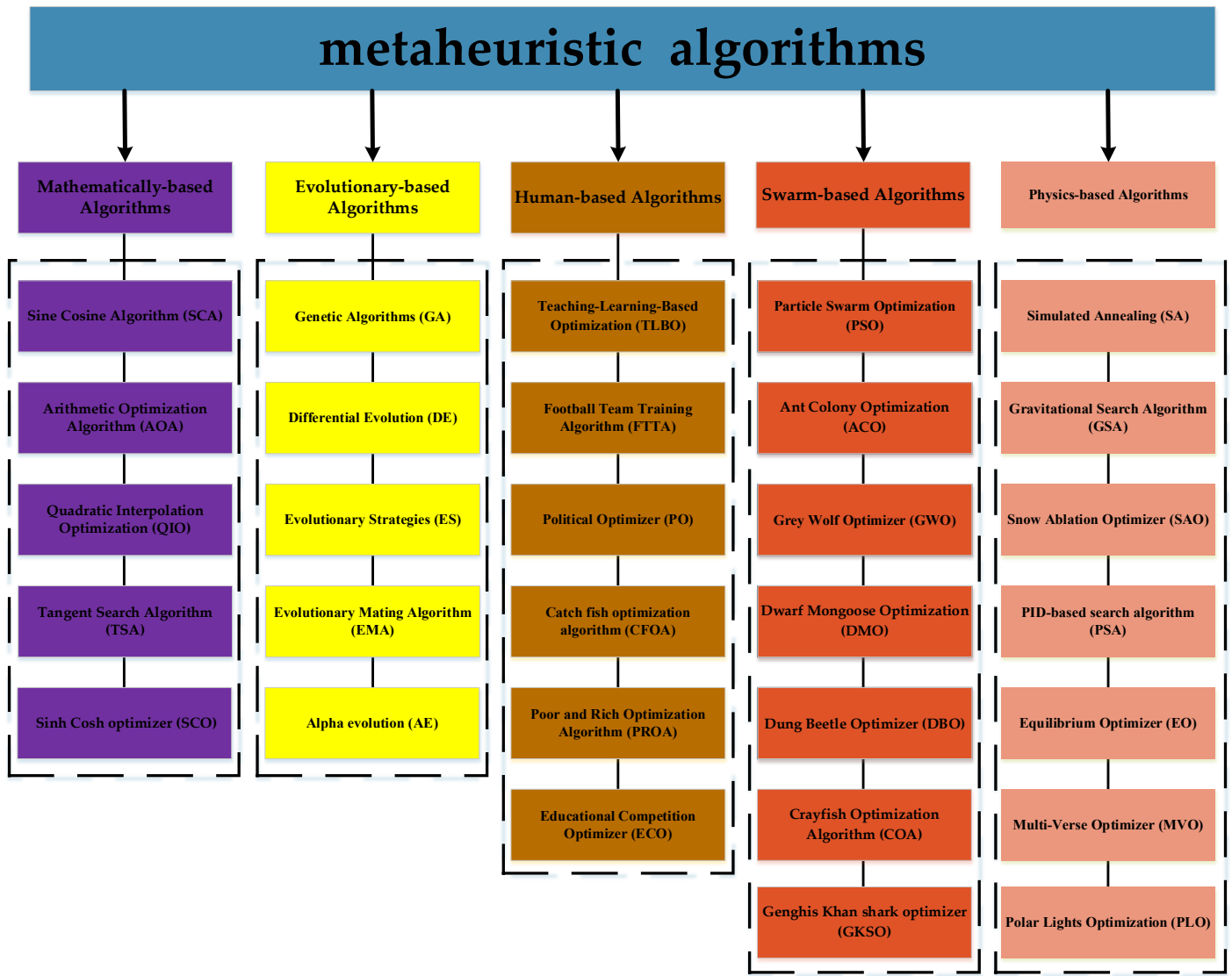

**Figure 1.** Classification of Metaheuristic Algorithm

Evolutionary-based algorithms primarily emulate biological phenomena like Darwin's evolution, genetic inheritance, and mating. These algorithms are rooted in the mechanisms of natural evolution, such as selection, mutation, and crossover, to simulate the survival of the fittest. Some of the well-known algorithms in this category include Genetic Algorithm (GA) [26], Differential Evolution (DE) [27], Evolutionary Strategies (ES) [28], Alpha Evolution (AE) [29], and Evolutionary Mating Algorithm (EMA) [30]. Swarm-based algorithms simulate the collective social behavior of foraging, reproduction, and avoidance of natural enemies in groups of organisms. Particle Swarm Optimization (PSO) [31] is one of the classical instances in which the foraging behavioral patterns of bird flocks are simulated to locate the ideal solution. Other prominent algorithms in this category comprise Ant Colony Optimization (ACO) [32], Grey Wolf Optimizer (GWO) [33], Dwarf Mongoose Optimization (DMO) [34], Hippopotamus Optimization Algorithm (HOA) [35], Crayfish Optimization Algorithm (COA) [36], Genghis Khan Shark Optimizer (GKSO) [37], Prairie Dog Optimization Algorithm (PDOA) [38], Artificial Lemming Algorithm (ALA) [39], Puma Optimizer (PO) [40], and Gazelle Optimization Algorithm (GOA) [41]. The third category comprises mathematical based algorithms. Algorithms based on mathematics are inspired by mathematical theories, functions, and formulas, which have demonstrated significant promise in enhancing the computing efficacy of optimization approaches. One of the noteworthy methods in this category is the Sine Cosine Algorithm (SCA) [42], which applies the concept of trigonometric functions to create an algorithmic model. Arithmetic Optimization Algorithm (AOA) [43], Quadratic Interpolation Optimization (QIO) [44], Tangent Search Algorithm (TSA) [45], and Sinh Cosh optimizer (SCO) [46] are other instances in this category. The fourth group encompasses physics-based algorithms that mimic the principles of physical phenomena in nature. One of the exemptional physical-based approaches is Simulated Annealing (SA) [47], which is influenced by the metallurgical annealing principle. Other meta-heuristic algorithms in this category are Gravitational Search Algorithm (GSA) [48], Snow Ablation Optimizer (SAO) [49], PID-based search algorithm (PSA) [50],

Equilibrium Optimizer (EO) [51], Multi-Verse Optimizer (MVO) [52], and Polar Lights Optimization (PLO) [53]. The final group involved human-based algorithms that are inspired by human social behaviors and collaborative strategies; these algorithms mimic human decision-making processes. Representative algorithms are the Teaching Learning Based Optimization (TLBO) [54], Football Team Training Algorithm (FTTA) [55], Political Optimizer (PO) [56], Catch Fish Optimization Algorithm (CFOA) [57], Poor and Rich Optimization Algorithm (PROA) [58], Hybrid Leader Based Optimization (HLBO) [59] and Educational Competition Optimizer (ECO) [60].

The Educational Competition Optimizer (ECO) was proposed in 2024 by Lian et al. as a novel metaheuristic algorithm inspired by the dynamics of social education. This algorithm divides the population into two groups: school agents and student agents, and employs a roulette wheel-based iterative framework to guide the search process. By utilizing progressive convergence and proximity-based selection strategies, ECO effectively reduces the risk of falling into local optima, thereby improving search performance compared to traditional optimization methods. In its initial study, ECO demonstrated strong convergence and exceptional search capabilities, achieving optimal performance on the CEC 2021 benchmark suite. However, the No Free Lunch (NFL) theorem suggests that this success may not generalize to other test problems. In the field of engineering, the solution of numerous practical problems can be attributed to optimization problems, but some existing optimization algorithms have many limitations in dealing with complex engineering scenarios. Although conventional meta-heuristic algorithms have a certain degree of versatility, they are still deficient in balancing exploration and exploitation, which leads to the tendency to fall into local optimums, slow convergence, and the accuracy of the solutions being improved when solving large-scale, high-dimensional, and multi-constraint engineering optimization problems. The basic ECO algorithms still have room for improvement in the balance between global search capability and local development capability, especially when facing high complexity engineering optimization problems, the quality of their convergence speed and solution may not be able to meet the practical demands [61, 62]. Considering the above, this paper proposes an improved educational competitive optimizer algorithm, with a view to overcoming the deficiencies of the existing algorithms in solving complex engineering optimization problems, and providing more efficient and accurate optimization solutions for engineering problems. At present, there are two classes of methods to improve the performance of the algorithms. The first category is to improve the initialization, including good point set initialization strategy [63, 64], chaotic mapping initialization strategy [65–67], and reverse learning initialization strategy [68–70], etc. The second category is to improve the search operator. This category can be used to improve the algorithm's ability to get rid of local optima by introducing the Levy or Weibull flight strategy [71–74]. The advantages of each algorithm can also be utilized by hybridizing different algorithms [75–81], etc. An improved ECO algorithm with hybrid mode search was proposed and used to optimize the PID controller by Serdar et al [62]. Marwa et al. strengthened the optimization capability of the ECO algorithm by combining a local escape operator and a Gaussian distribution strategy, and also solved the energy optimization problem [82]. Tang et al. utilized a fitness distance balancing strategy to select the optimal individual to achieve a balance between ECO exploration and exploitation [61]. Oluwatayomi et al. proposed an improved artificial electric field algorithm combining Gaussian variational specular reflection learning and a local escape operator, which increases the convergence speed and the ability to avoid falling into the local optimum point of a given problem [83].

To address these challenges, this paper introduces multi-covariance learning operators to refine the basic ECO algorithm, aiming to enhance both exploration and exploitation of the search space, thereby improving solution quality and convergence speed. This study proposes a novel enhanced variant of ECO based on multi-covariance learning operators, termed IECO-MCO. IECO-MCO integrates three distinct covariance learning operators to diversify and deepen the search process. The Gaussian covariance learning operator emphasizes broad exploration, the shift covariance learning operator focuses on balancing global exploration and local exploitation, and the differential covariance learning operator aims to enrich population diversity and prevent the algorithm from converging to local optima. The effectiveness of IECO is evaluated using benchmark functions from CEC 2017 and CEC 2022, and its performance is compared with various basic and improved algorithms across different categories. The comparative analysis includes: (1) ablation algorithms—ECO with only the Gaussian covariance operator (GECO), ECO with only the shift covariance operator (SECO), ECO with only the differential covariance operator (DECO), and the basic ECO; (2) basic algorithms: SAO, CFOA, DBO, QIO, and AE; and (3) improved algorithms: RDGMVO, ISGTOA, AFDBARO, MTVSCA, and ALSHADE. The results reveal that the IECO algorithm surpasses the basic ECO algorithm and other competitors in convergence speed, stability, and the capability to avoid local optima. Furthermore, statistical analyses, including the Friedman test, Wilcoxon rank-sum test, and Kruskal-Wallis test, are conducted to validate the superiority of IECO over the compared algorithms. Additionally, the practical applicability of the proposed IECO algorithm is verified through addressing constrained optimization challenges in engineering design. The experimental results highlight the exceptional performance of IECO in addressing complex op-timization challenges, underscoring its robustness and effectiveness in real-world applications. This study makes significant contributions in several key areas:

(a) By integrating three distinct covariance learning operators into the basic ECO algorithm, an upgraded version, IECO-MCO, is proposed. The Gaussian covariance operator is used to replace the school population at the primary stage. The shift covariance operator is used to replace the school population at the middle school stage. The differential covariance operator is used to update the school population at the high school stage.

(b) Three covariance learning operators are proposed. Gaussian Covariance Operator: This operator guides the population toward advantageous directions, thereby improving the overall population quality. By promoting broader exploration, it enhances search efficiency, enabling the algorithm to better understand the global landscape of the search space. Shift Covariance Operator: This operator adjusts the movement direction of elite agents using multiple reference points, ensuring that each agent follows a distinct trajectory. This approach achieves an appropriate balance between exploration and exploitation. Differential Covariance Operator: By leveraging the differences between random solutions, this operator maintains population diversity during the optimization process, thereby enhancing the robustness of the ECO framework.

(c) The efficacy of IECO-MCO is rigorously examined using the CEC2017 and CEC2022 benchmark test suites. The assessment incorporates ablation experiment analysis and statistical analyses employing the Friedman test, Kruskal-Wallis test, and Wilcoxon rank sum test.

(d) IECO-MCO is successfully applied to engineering design optimization problems, demonstrating its robustness and adaptability in practical applications.

(e) IECO-MCO proves to be competitive against a wide range of basic and improved algorithms from different categories, such as SAO, CFOA, DBO, QIO, AE, RDGMVO, ISGTOA, AFDBARO, MTVSCA, and ALSHADE, showcasing its cutting-edge capabilities.

The organization of this paper is outlined as follows: Section 2 offers an overview of the foundational principles and mathematical formulation underlying the original ECO algorithm. Section 3 elaborates on the core ideas behind the proposed IECO-MCO methodology, accompanied by detailed pseudo-code, flowcharts, and an in-depth analysis of its computational complexity. Section 4 evaluates the optimization capabilities of IECO-MCO using the CEC2017 and CEC2022 benchmark suites, while also examining the contributions of each enhancement strategy to its overall performance. Section 5 thoroughly evaluates the proposed IECO-MCO algorithm on engineering design problems. Finally, Section 6 summarizes the principal findings of this research and highlights potential directions for future research. A comprehensive schematic of the paper's organization is presented in Figure 2.

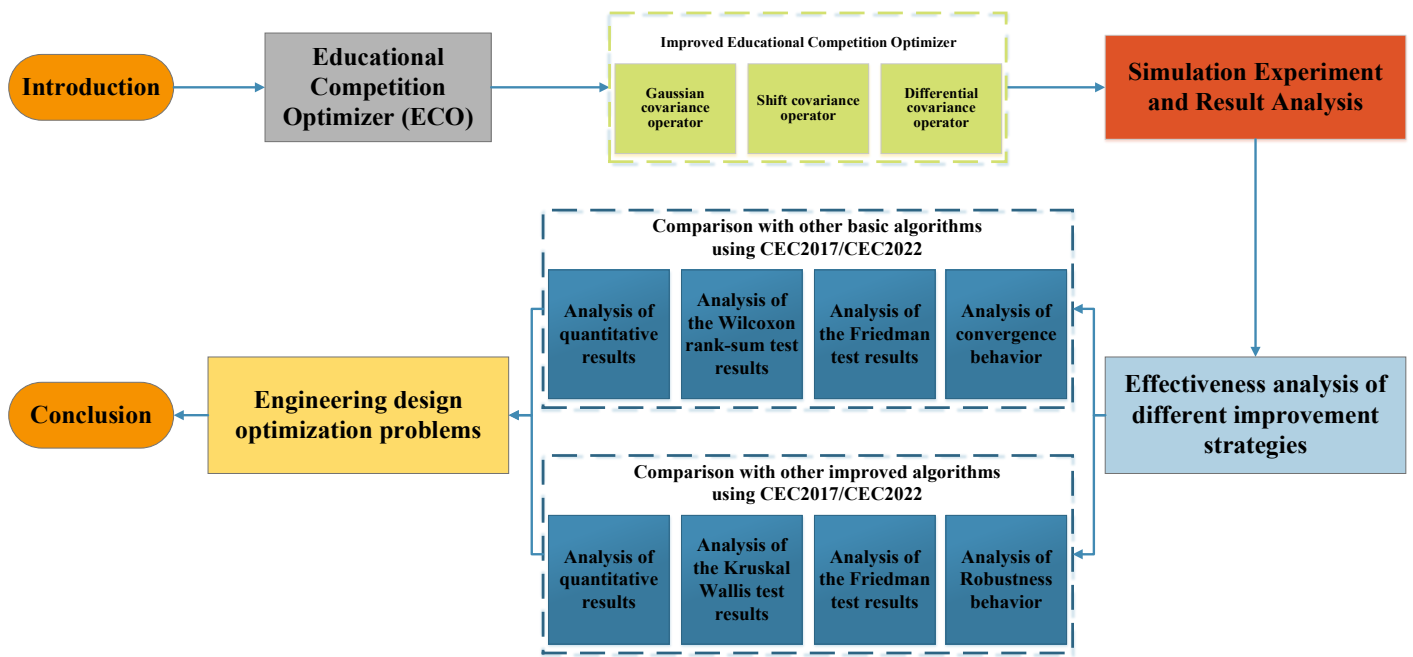

**Figure 2.** A complete outline of this study.

## 2. Educational Competition Optimizer (ECO)

The ECO algorithm is designed to simulate the competitive dynamics inherent in educational systems, reflecting the evolving strategies seen across primary, middle, and high school levels. As competitive pressures increase and the number of available schools declines, the optimization process of the ECO algorithm is structured into three sequential stages. These stages facilitate a smooth transition from broad exploration to focused exploitation, utilizing an adaptive search approach. Within a mathematical optimization framework, the algorithm seeks to derive optimal solutions under defined constraints. The mathematical representation of the ECO algorithm is outlined as follows.

### 2.1. Population initialization

The population initialization process of the ECO algorithm distinguishes itself from other metaheuristic algorithms by employing logistic chaotic mapping to generate the initial population. This approach is designed to simulate the social disorder phenomenon resulting from a lack of education. The initialization formula for logistic chaos mapping technology, given a population size of $N$, and problem space boundaries defined by the $Lb$ (lower bound) and $Ub$ (upper bound), can be expressed as Equation 1:

$$x_i = \alpha \times x_{i-1} \times (1 - x_{i-1}), 0 \leq x_0 \leq 1, i = 1, 2, ..., N \tag{1}$$

where $x_i$ represents the $i^{th}$ value and $x_{i-1}$ represents the previous value. The parameter $\alpha$ is defined as a constant, with its value set to 4 in this study. Map the chaotic value $x_i$ to the search space using Equation 2:

$$X_i = Lb + (Ub - Lb) \times x_i \tag{2}$$

### 2.2. Primary school stage

In the primary school phase, the ECO algorithm segregates the population into two distinct subpopulations. During each iteration, the top 20% of individuals, ranked according to their fitness values, are designated as school agents, while the remaining 80% are assigned as student agents. School agents establish their optimal positions by analyzing the average location of the population. Simultaneously, student agents determine their attending schools based on their distance to nearby schools. This approach embodies the initial exploration phase, where schools and students explore potential locations with limited constraints. The mathematical formulations describing the behaviors of school agents and student agents are provided in Equation 3 and Equation 4.

$$School: X_i^{new} = X_i^{now} + \omega \times (X_{imean}^{now} - X_i^{now}) \times Levy(D) \tag{3}$$

$$Students: X_i^{new} = X_i^{now} + \omega \times (close(X_i^{now}) - X_i^{now}) \times randn \tag{4}$$

$$\omega = 0.1 \times \ln\left(2 - \frac{FEs}{FEs_{max}}\right) \tag{5}$$

where $X_i^{now}$ represents the current agent's position. $X_i^{new}$ denotes the position after updating. $X_{imean}^{now}$ represents the average position of the vector for the $i^{th}$ school agent. $Levy(D)$ is a random vector that follows the Lévy distribution. $close(X_i^{now})$ indicates the location of the school closest to $X_i^{now}$. $randn$ represents a random variable following a normal distribution. $FEs$ and $FEs_{max}$ represent the current number of function evaluations and the maximum number of function evaluations, respectively.

### 2.3. Middle school stage

During the middle school phase, the ECO algorithm continues to partition the population into school agents and student agents. The school agents comprise the top 10% of individuals ranked by their fitness values, while the remaining 90% are classified as student agents. School agents employ a more advanced strategy to determine their locations, incorporating both the average position of the agents and the position of the best agent. Similarly, student agents in this phase select their attending schools based on their proximity to nearby schools. Furthermore, student agents are subdivided into two groups depending on their academic potential. The mathematical expressions describing the behaviors of school agents and student agents are detailed in Equation 6 and Equation 7.

$$School: X_i^{new} = X_i^{now} + \left(X_{best}^{now} - X_{mean}^{now}\right) \times e^{\left(\frac{FEs}{FEs_{max}} - 1\right)} \times Levy(D) \tag{6}$$

$$Students: X_i^{new} = X_i^{now} - \omega \times close\left(X_i^{now}\right) - P \times \left(E \times \omega \times close\left(X_i^{now}\right) - X_i^{now}\right) \tag{7}$$

$$P = 4 \times randn \times \left(1 - \frac{FEs}{FEs_{max}}\right) \tag{8}$$

$$E = \begin{cases} \frac{\pi}{P} \times \frac{FEs}{FEs_{max}}, & R_m > Th \\ 1, R_m \leq Th \end{cases} \tag{9}$$

where $X_{best}^{now}$ denotes the position of the best school agent. $X_{mean}^{now}$ represents the average position of whole agents. The talent values of different students are simulated using the random number $R_m$, which takes on a value within the range of [0, 1]. The parameter $Th$ is set to 0.5.

*2.4. High school stage*

In the high school phase, the ECO algorithm maintains the division of the population into school agents and student agents, with the proportions of each subpopulation remaining consistent with the middle school phase. School agents adopt a more cautious approach when selecting their locations, taking into account not only the average position of the agents but also the best and worst position within the population. This comprehensive evaluation enables schools to make informed decisions about their locations, thereby addressing the broader needs of the student population. In contrast, student agents choose to attend the currently best-performing school. The mathematical expressions describing the behaviors of school agents and student agents are detailed in Equation 10 and Equation 11.

$$School: X_i^{new} = X_i^{now} + \left(X_{best}^{now} - X_{mean}^{now}\right) \times randn - \left(X_{worst}^{now} - X_{mean}^{now}\right) \times randn \tag{10}$$

$$Students: X_i^{new} = X_i^{now} - P \times \left(E \times X_{best}^{now} - X_i^{now}\right) \tag{11}$$

$$E = \begin{cases} \frac{\pi}{P} \times \frac{FEs}{FEs_{max}}, & R_h > Th \\ 1, R_h \leq Th \end{cases} \tag{12}$$

where $X_{worst}^{now}$ represents position of the worst school agent. The talents of individual students are represented by a random number denoted as $R_h$, which falls within the range of [0, 1].

## 3. Improved Educational Competition Optimizer

While the Educational Competition Optimizer (ECO) demonstrates promise as a metaheuristic algorithm, it faces notable challenges in complex optimization scenarios. ECO exhibits inherent limitations, including difficulties in balancing exploitation and exploration, a tendency to converge to local optima, and suboptimal convergence accuracy. These weaknesses hinder its effectiveness in tackling intricate and demanding optimization problems. To tackle these limitations, this paper proposes the integration of three distinct covariance learning operators. These operators are designed to achieve a robust equilibrium between exploitation and exploration, promote extensive information exchange within the population, and ultimately enhance the algorithm's optimization performance. In this section, the structure of the proposed IECO-MCO algorithm is initially detailed. Subsequently, the pseudo-code and flowchart are presented. Finally, this section performs an analysis of the computational complexity related to the IECO-MCO algorithm introduced in this study.

*3.1. Multi-covariance learning operators*

In the covariance learning strategy, the Gaussian probability distribution model is employed to characterize the distribution of high-performing agents, thereby providing a directional guide for population evolution. This approach is mainly structured into two stages: sampling and generation. In the sampling stage, the maximum likelihood estimation technique is employed to approximate the Gaussian distribution. The covariance matrix $C$ and the weighted average position of the high-performing agents $X_{mean}^{better}$ can be derived using Equation 13 and Equation 14, respectively.

$$C = \frac{1}{|P_d|} \sum_{i=1}^{|P_d|} \left(X_i^S - X_{mean}^{better}\right) \times \left(X_i^S - X_{mean}^{better}\right)^T \tag{13}$$

$$X_{mean}^{better} = \sum_{i=1}^{|S|} \omega_i \times X_i^S \tag{14}$$

$$\omega_i = \ln\left(|P_d|+1\right) / \left(\sum_{i=1}^{|P_d|}\left(\ln\frac{\left(|P_d|+1\right)}{i}\right)\right) \tag{15}$$

where $P_d$ denotes the ensemble of dominant groups and $|P_d|$ denotes the number of dominant groups. $\omega_i$ is the weight coefficient. $X_i^S$ denotes the agent of dominant groups. By introducing weighting coefficients, it is ensured that the better individuals have a greater impact on population evolution while retaining effective information from more individuals. After the sampling phase is concluded, offspring individuals are produced using the following three methods.

Gaussian covariance operator: In the primary school phase, the school population adjusts its position based solely on its own location, overlooking valuable information from superior groups, which hinders comprehensive exploration. To address this limitation, this paper introduces the Gaussian covariance operator to replace the original position update method for the school population. The Gaussian covariance operator integrates information from high-performing populations and accounts for the differences between individual and collective data, thereby enhancing the algorithm's global exploration capabilities. The Gaussian covariance operator is computed using Equation 16.

$$X_i^{new} = Gaussian\left(X_{mean}^{better}, C\right) + rand \times \left(X_{mean}^{better} - X_i^{now}\right) \tag{16}$$

Shift covariance operator: In the middle school phase, the school population adjusts its position based on the average location of the population and the position of the best-performing agent. While considering the average position of all agents may weaken the algorithm's convergence ability, relying solely on the optimal agent's position elevates the probability of becoming trapped in local optima. The weighted average position $X_{mean}^{better}$ plays a crucial role in influencing population distribution. To tackle these challenges, this paper introduces a shift covariance operator to replace the position update method for the school population during the middle school phase. The shift covariance operator is designed to maintain a balance between exploration and exploitation, thereby preventing premature convergence to local optima. This strategy integrates information from the current best position $X_{best}^{now}$, the weighted position of high-performing groups $X_{mean}^{better}$, and individual position $X_i^{now}$, achieving a distribution with varying means. This approach enhances population diversity and improves exploration performance. The shift covariance operator is computed using Equation 17.

$$X_i^{new} = Gaussian\left(\left(\frac{X_{mean}^{better} + X_{best}^{now} + X_i^{now}}{3}\right), C\right) + rand \times \left(X_{mean}^{better} - X_i^{now}\right) \tag{17}$$

Differential covariance operator: In the high school phase, the school population in ECO adjusts its position based on the differences between itself, the best-performing individual, and the worst-performing individual. This approach fails to incorporate information from the entire population, often leading to reduced diversity and a tendency to converge to local optima. To address this limitation, this paper employs the differential covariance operator to replace the position update method for the school population during the high school phase. This operator significantly enhances population diversity and aids the algorithm in escaping local optima. It achieves this by utilizing random solutions $X_{ran1}^{now}$ and $X_{ran2}^{now}$, weighted solutions from high-performing groups

$X_{mean}^{better}$, the best solution $X_{best}^{now}$, and the worst solution $X_{worst}^{now}$ during the search process. The incorporation of random solutions during the mutation phase facilitates exploration of diverse regions within the search space. The school population moves in the direction of a random vector, guided by the weighted position of high-performing groups, while incorporating the differences between two randomly selected solutions and the best and worst solutions. The differential covariance operator is computed using Equation 18.

$$X_i^{new} = Gaussian\left(X_{mean}^{better}, C\right) + rand \times \left(X_{ran1}^{now} - X_{best}^{now}\right) + rand \times \left(X_{ran2}^{now} - X_{worst}^{now}\right) \tag{18}$$

As shown in Equation 13, the selection of elite individuals plays a pivotal role in the performance of the covariance learning strategy. In the CMA-ES algorithm, elite individuals are chosen based solely on fitness values. However, since the Gaussian probability distribution model simulates the distribution of certain individuals, relying exclusively on fitness for selection can lead to overfitting of the distribution model. To address this issue, this paper introduces a combined score of fitness and distance as the selection criterion. Agents with high fitness may not necessarily be concentrated in distribution, and those concentrated in distribution may not always exhibit superior fitness. By integrating these two metrics for selecting elite individuals, the sampling process becomes more selective, thereby enhancing the diversity of the elite population. Furthermore, as the optimization progresses, the number of retained elite individuals increases over multiple iterations, which may result in the overuse of older individuals and stagnation in evolution. To mitigate this, a first-in-first-out approach is adopted, where newly selected individuals replace the earliest retained elite individuals when the population size exceeds a threshold.

*3.2. The framework of IECO-MCO*

The enhanced IECO-MCO algorithm is developed by integrating the foundational ECO framework with multi-covariance learning operators outlined earlier. To provide a clear understanding of the proposed IECO-MCO, the corresponding pseudo-code is detailed in Algorithm 1. Additionally, Figure 3 offers a comprehensive flowchart that delineates the operational steps of the IECO-MCO algorithm.

**Algorithm 1**: Pseudo-code of the IECO-MCO

1: Initialize the ECO parameters
2: Initialize the solutions' positions randomly, Eq. (2)
3: **While** $Fes < FEs_{max}$
4: Calculate the fitness function
5: Find the best position and worst position
6: Calculate $R_m$, $R_p$, $P$, $E$, $C$
7: **For** $i = 1: N$ **do**
8:   Stage 1: Primary school competition
9:     **If** mod ($Fes$, 3) == 1 **Then**
10:       Update school position by Eq. (16)       // Gaussian covariance operator
11:       Update student position by Eq. (4)
12:     **End if**
13:   Stage 2: middle school competition
14:     **If** mod ($Fes$, 3) == 2 **Then**
15:       Update school position by Eq. (17)       // Shift covariance operator
16:       Update student position by Eq. (7)
17:     **End if**
18:   Stage 3: High school competition
19:     **If** mod ($Fes$, 3) == 3 **Then**
20:       Update school position by Eq. (18)       // Differential covariance operator

21:     Update student position by Eq. (11)
22:     **End if**
23: **End for**
24: *FEs = Fes + N*
24: **End while**
25: Return the best solution

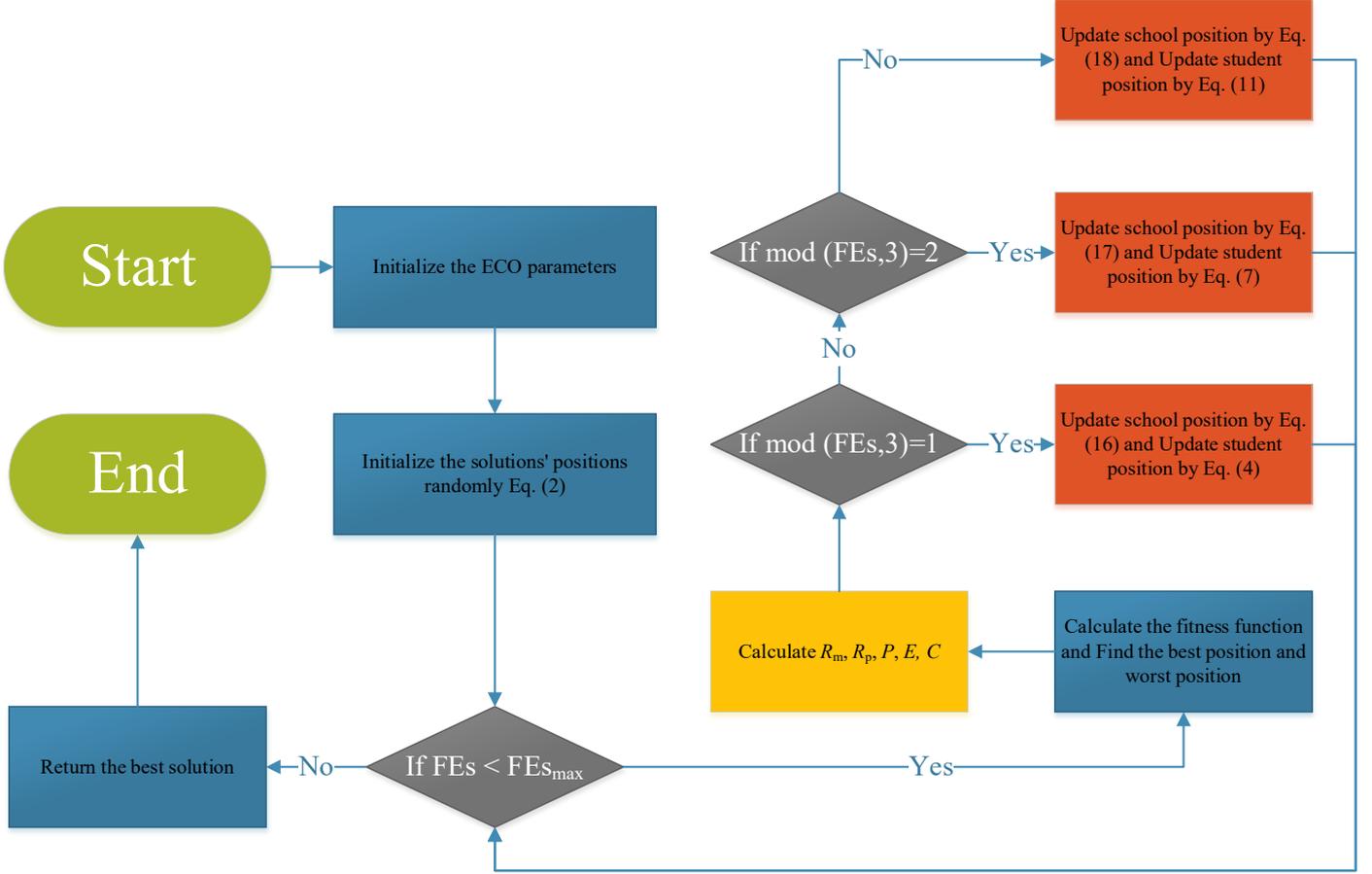

**Figure 3.** The flowchart of the suggested IECO-MCO algorithm

*3.3. Complexity analysis of IECO-MCO*

Complexity analysis serves as a holistic evaluation of a system, assessing the efficiency of an algorithm. Let $N$ indicates the size of the population, $T$ denote the maximum number of iterations, and $D$ signify the problem's dimensionality. The computational complexity of the basic ECO algorithm is $O(T \times N \times D)$.

For IECO-MCO, the introduction of the three proposed covariance learning operators replaces the original position update method for the school population without increasing its time complexity. The time complexity of the student population remains unchanged. Consequently, the time complexity of the proposed IECO-MCO is $O(T \times N \times D)$, identical to that of the basic ECO algorithm, ensuring no additional computational overhead is incurred.

## 4. Simulation Experiment and Result Analysis

To evaluate the optimization performance of the proposed IECO-MCO, this section provides a detailed report on its performance across 29 functions from the CEC2017 benchmark suite and 12 functions from the CEC2022 benchmark suite. To comprehensively demonstrate the capabilities of IECO-MCO, the selected comparison algorithms encompass a variety of basic and improved methods from different categories:

(1) Evolution-based algorithms: AE [29] and ALSHADE [84];
(2) Physics-based algorithms: SAO [49] and RDGMVO [85];

(3) Swarm-based algorithms: DBO [86] and AFDBARO [87];
(4) Mathematics-based algorithms: QIO [44] and MTVSCA [88];
(5) Human-based algorithms: CFOA [57] and ISGTOA [89].

To ensure a fair comparison and account for the stochastic nature of the algorithms, each algorithm is independently executed 30 times for every function, with the maximum number of function evaluations set to 3000×$D$, where $D$ represents the problem dimensionality. All experiment evaluations in this research are performed using an identical platform, with specifications detailed in Table 1. Furthermore, the parameters for all competing algorithms are configured according to their original publications, as summarized in Table 2.

**Table 1.** Platform specifications utilized for experiments

| Hardware | |
|---|---|
| CPU | AMD R9 7945HX |
| RAM | 32 GB |
| Operating system | 64-bit Windows 11 |
| Platform type | R9000P 2024 |
| **Software** | |
| Programming language | MATLAB R2023a |

**Table 2.** Parameter settings of MRIME-CD and other competing algorithms.

| Common setting | |
|---|---|
| Maximum number of function evaluations | $FEs_{max} = 3000 \times D$ |
| Dimension of the Problem | $D = 10/30/50/100 \, (\text{CEC2017})$ <br> $D = 10/20 \, (\text{CEC2022})$ |
| Number of independent runs | 30 |
| **Algorithm setting** | |
| IECO-MCO | $H = 0.5, G_1 = 0.4, G_2 = 0.5, |S| = 20D$ |
| ECO | $H = 0.5, G_1 = 0.2, G_2 = 0.1$ |
| SAO | $k = 1$ |
| CFOA | No parameter |
| DBO | $p = 0.2, r = 0.9$ |
| QIO | $a_1 = 0.7, a_2 = 0.15, a_3 = 3, b = 2$ |
| AE | $\omega = 2, S = rand < 0.5, l = 0.5$ |
| RDGMVO | $W_{max} = 1, W_{min} = 0.2, p = 0.6$ |
| ISGTOA | $Nm = 2$ |
| AFDBARO | $p = 1, \omega = 0.5$ |
| MTVSCA | $\lambda = 0.25, c = 0.7, a = 2$ |
| ALSHADE | $F = 0.5, CR = 0.5, p = 0.11, H = 6, P = 0.5$ |

The efficacy of an algorithm is typically affected by multiple factors. A key method for assessing its efficacy involves analyzing the outcomes generated by applying the proposed approach to diverse problem categories. Within this framework, this section seeks to evaluate the feasibility, applicability, and reliability of the proposed IECO-MCO algorithm. To this end, two benchmark test suites from the IEEE Congress on Evolutionary Computation are employed to thoroughly analyze IECO-MCO. These functions are widely regarded as nonlinear, non-differentiable, non-convex, and highly intricate. A detailed summary of these test functions is presented in Table 3.

Table 3. Detailed description of CEC2017 and CEC2022 test functions

| Test suite | Type | ID |
|---|---|---|
| CEC2017 | Unimodal functions | F1-F2 |
| | Multimodal functions | F3-F9 |
| | Hybrid functions | F10-F19 |
| | Composition functions | F20-F29 |
| CEC2022 | Unimodal functions | F1 |
| | Basic functions | F2-F5 |
| | Hybrid functions | F6-F8 |
| | Composition functions | F9-F12 |

*4.2. Effectiveness analysis of different improvement strategies*

This paper proposes three covariance learning operators to augment the performance of the fundamental ECO algorithm, necessitating an evaluation of each operator's contribution to the efficacy of IECO-MCO. To comprehensively demonstrate the influence of each operator on the test functions, the CEC2017 benchmark suite with dimensions of 10, 30, 50, and 100, as well as the CEC2022 benchmark suite with dimensions of 10 and 20, are employed to assess the effectiveness of these enhancement strategies. In these experiments, IECO-MCO is compared with three variants and the baseline algorithm: (1) GECO: incorporating solely the Gaussian covariance operator, (2) SECO: incorporating solely the shift covariance operator, (3) DECO: incorporating solely the differential covariance operator, and (4) ECO: serving as the baseline for comparison. The performance of these variants elucidates the positive impact of distinct operators on the ECO algorithm. Moreover, by juxtaposing IECO-MCO with these variants, the synergy and compatibility of the three operators can be validated.

Given the extensive volume of experimental data, presenting detailed results may hinder readability. Therefore, this paper employs three statistical tests to analyze the experimental data. Table 6 summarizes the Friedman test results for IECO-MCO and its variants. The Friedman test, also referred to as the multiple comparison test, is utilized to identify significant differences among multiple algorithmic approaches. Based on the p-values from the Friedman test in Table 4, it is evident that significant differences exist among these algorithms. The proposed IECO-MCO demonstrates the best overall performance, achieving average ranks of 1.93 and 1.79 on the two test sets, respectively. In contrast, the basic ECO ranks last in both tests, with average ranks of 4.26 and 4.38.

Table 4. The Friedman ranking of IECO-MCO and its variants

| Test suite | Dimension | ECO | GECO | SECO | DECO | IECO-MCO | P-value |
|---|---|---|---|---|---|---|---|
| CEC 2017 | 10 | 4.28 | 3.38 | 3.17 | 2.28 | **1.90** | 2.56E-08 |
| | 30 | 4.66 | 2.66 | 3.83 | 2.45 | **1.41** | 3.54E-15 |
| | 50 | 4.17 | 2.69 | 3.45 | 2.72 | **1.97** | 1.38E-06 |
| | 100 | 3.93 | 2.48 | 4.10 | **2.03** | 2.45 | 1.99E-08 |
| Mean rank | | 4.26 | 2.80 | 3.64 | 2.37 | **1.93** | N/A |
| CEC 2022 | 10 | 4.58 | 2.92 | 2.92 | 2.42 | **2.17** | 1.88E-03 |
| | 20 | 4.67 | 2.75 | 3.50 | 2.67 | **1.42** | 1.65E-05 |
| Mean rank | | 4.63 | 2.83 | 3.21 | 2.54 | **1.79** | N/A |
| Total mean rank | | 4.38 | 2.81 | 3.49 | 2.43 | **1.88** | N/A |

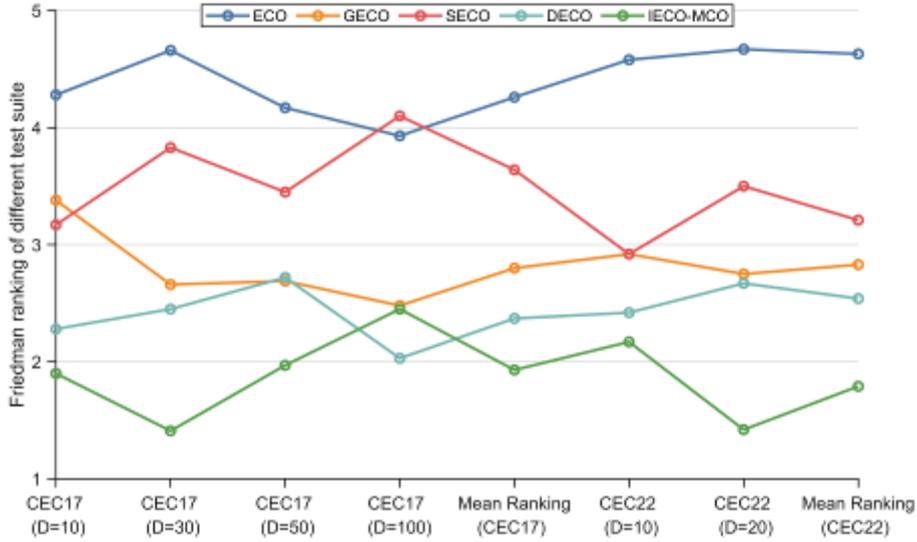

**Figure 4.** The visualization of the Friedman ranking of IECO-MCO and its variants

Figure 4 presents a line graph based on the Friedman test rankings of these algorithms, illustrating the trend of each algorithm's ranking. The performance of IECO-MCO remains stable across different dimensions within the same test set, with minimal variation across different test sets, clearly demonstrating its superior performance and stability regardless of dimensionality. Variants of ECO incorporating a single covariance learning operator also outperform the basic ECO algorithm, indicating that all three covariance learning operators enhance ECO's performance. Specifically, the differential covariance operator has the most significant impact on IECO-MCO, followed by the shift covariance operator and the Gaussian covariance operator. Additionally, although IECO-MCO underperforms DECO on CEC2017 100D, it exhibits better overall performance and outperforms DECO in all other scenarios.

**Table 5.** The Wilcoxon rank sum test results (+/ = /-) of IECO-MCO and its variants

| vs. ECO +/=/- | CEC-2017 test suite | | | | CEC-2022 test suite | |
|---|---|---|---|---|---|---|
| | 10D | 30D | 50D | 100D | 10D | 20D |
| GECO | 15/13/1 | 23/5/1 | 23/5/1 | 24/2/3 | 7/5/0 | 9/3/0 |
| SECO | 18/11/0 | 13/14/2 | 10/15/4 | 8/10/11 | 7/5/0 | 9/3/0 |
| DECO | 19/10/0 | 20/8/1 | 16/12/1 | 25/3/1 | 7/5/0 | 7/4/1 |
| IECO-MCO | 22/4/3 | 26/2/1 | 22/4/3 | 20/4/5 | 10/2/0 | 11/1/0 |

Table 5 summarizes the Wilcoxon rank-sum test results for IECO-MCO, ECO variants incorporating different covariance learning operators, and the basic ECO at a 5% significance level. The non-parametric Wilcoxon test enables the analysis of significant differences between IECO-MCO and other variants. The symbols "+/=/−" in Table 5 indicate that IECO-MCO performs "better," "similar," or "worse" compared to other algorithms, respectively. According to Table 5, except for SECO's results on the CEC2017 100D function, all improved algorithms achieve more "+" than "−" in other cases, demonstrating that IECO-MCO and the three variants generally outperform the basic ECO algorithm. GECO obtains more "+" than IECO-MCO when handling the CEC2017 50D and 100D functions, while DECO achieves more "+" than IECO-MCO on the CEC2017 100D function. However, IECO-MCO secures the most "+" in other scenarios. Further analysis reveals that the three learning operators synergistically enhance ECO's performance on low-dimensional problems, converting more "similar" (=) cases into "superior" (+) outcomes. For high-dimensional problems, the Gaussian covariance operator and differential covariance operator contribute more significantly to ECO's improvement. The results of the Wilcoxon rank-sum test align with those of the Friedman test, further validating the effectiveness of the proposed enhancement strategies.

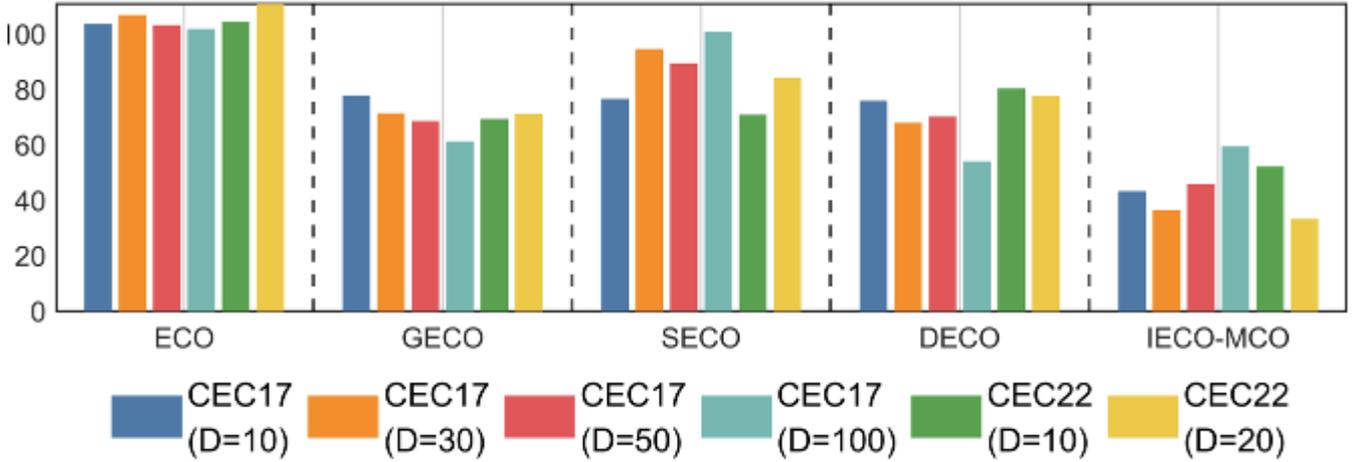

**Figure 5.** Kruskal-Wallis ranking of IECO-MCO and its variants

Additionally, the Kruskal-Wallis test is employed to further analyze the differences between IECO-MCO, its derived variants, and the basic ECO algorithm. The Kruskal-Wallis test evaluates the overall ranking of different metaheuristic algorithms by comparing the average ranks of the algorithms. Figure 5 presents the Kruskal-Wallis test results across different dimensions, where the smallest rank value indicates the superior performance of the corresponding algorithm. According to Figure 5, IECO-MCO outperforms other algorithms in all cases except for CEC2017 100D, providing lower rank values. In conclusion, the three proposed enhancement strategies are complementary and work synergistically to improve the performance of IECO-MCO, making it adaptable to various types of optimization problems.

*4.3. Comparison with other basic algorithms using CEC2017/CEC2022*

In this subsection, alongside the basic ECO algorithm, several basic metaheuristic algorithms from diverse categories—SAO, CFOA, DBO, QIO, and AE—are incorporated into the experiments. The experimental environment and parameter configurations are outlined in Table 1 and Table 2. The detailed analysis results are presented below.

4.3.1. Analysis of quantitative results

Due to the extensive volume of data across six scenarios, only the rankings based on the "average values" of each algorithm for each test function are displayed here to facilitate reader comprehension. The best value, mean, and standard deviation are recorded in Table S1-Table S6 in the supplementary files. Figure 6 illustrates the rankings of each algorithm on every test function. By analyzing Figure 6, we can gain a clear understanding of the performance of IECO-MCO and these basic algorithms across the two test sets.

According to Figure 6, IECO-MCO ranks within the top three for the majority of the 29 functions in CEC2017 and the 12 functions in CEC2022, while the basic ECO predominantly ranks in the bottom three for most functions. IECO-MCO achieves first place in 14, 12, 11, and 10 instances across the four scenarios of CEC2017, and in 8 and 6 instances across the two scenarios of CEC2022. Although AE secures more first-place rankings in the 50D and 100D cases of CEC2017, its performance is less stable, exhibiting significant variability across different functions. In contrast, IECO-MCO demonstrates consistent performance across varying dimensions and functions, maintaining exceptional accuracy and robustness regardless of increases in dimensionality or changes in problem types. Therefore, we can conclude that the proposed IECO-MCO exhibits superior potential for solving complex optimization problems compared to the basic algorithms.

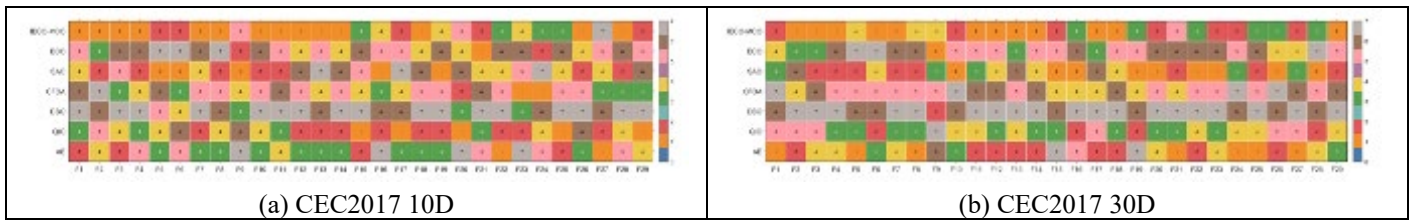

| (a) CEC2017 10D | (b) CEC2017 30D |

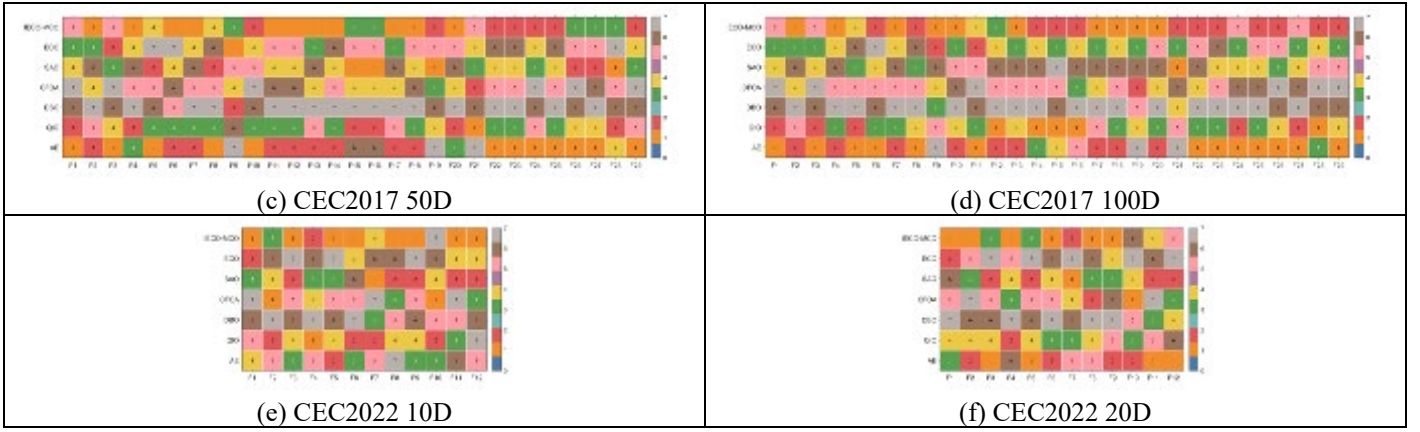

Figure 6. Ranking of IECO-MCO and other basic algorithms based on mean values

4.3.2. Analysis of the Wilcoxon rank-sum test results

To address the randomness inherent in experimental results, this section employs statistical tests based on multiple runs to further analyze the outcomes. The Wilcoxon rank-sum test is first utilized to determine the performance differences between IECO-MCO and other basic algorithms. Table 6 summarizes the Wilcoxon rank-sum test results for IECO-MCO and the basic competitors across the two test sets. Figure 7 illustrates the number of "+/=/−" obtained by IECO-MCO. According to the Wilcoxon rank-sum test results, when comparing IECO-MCO with AE on CEC2017 100D, IECO-MCO achieves fewer "+" than "−", indicating that IECO-MCO's performance on CEC2017 100D is inferior to that of AE. However, in the other five scenarios, IECO-MCO secures more "+" than "−", demonstrating that IECO-MCO significantly outperforms AE in overall performance. IECO-MCO exhibits slightly lower efficiency than SAO when tackling the CEC2022 20D problem, but the difference between the two is minimal. For other basic algorithms, IECO-MCO consistently maintains an advantage, achieving more "+" than "−". Notably, compared to ECO, MRIMR-CD attains "+" on most functions, highlighting significant differences between IECO-MCO and ECO in the majority of cases. The performance markers of IECO-MCO against different basic competitors are presented as follows.

For the CEC2017 10D functions, the performance of IECO-MCO surpasses (falls short of) ECO on 24(3) functions, SAO on 17(6) functions, CFOA on 22(3) functions, DBO on 23(2) functions, QIO on 17(8) functions, and AE on 21(3) functions.

For the CEC2017 30D functions, the performance of IECO-MCO surpasses (falls short of) ECO on 28(0) functions, SAO on 11(69) functions, CFOA on 27(1) functions, DBO on 28(0) functions, QIO on 21(2) functions, and AE on 19(7) functions.

For the CEC2017 50D functions, the performance of IECO-MCO surpasses (falls short of) ECO on 24(2) functions, SAO on 16(11) functions, CFOA on 27(1) functions, DBO on 27(0) functions, QIO on 19(6) functions, and AE on 15(12) functions.

For the CEC2017 100D functions, the performance of IECO-MCO surpasses (falls short of) ECO on 22(4) functions, SAO on 22(2) functions, CFOA on 29(0) functions, DBO on 27(0) functions, QIO on 16(9) functions, and AE on 11(13) functions.

For the CEC2022 10D functions, the performance of IECO-MCO surpasses (falls short of) ECO on 10(1) functions, SAO on 7(1) functions, CFOA on 9(1) functions, DBO on 8(1) functions, QIO on 8(1) functions, and AE on 9(1) functions.

For the CEC2022 20D functions, the performance of IECO-MCO surpasses (falls short of) ECO on 11(0) functions, SAO on 4(5) functions, CFOA on 9(0) functions, DBO on 9(2) functions, QIO on 9(1) functions, and AE on 7(5) functions.

Table 6. The Wilcoxon rank sum test results (+/ = /-) of IECO-MCO and other basic competitors

| IECO-MCO vs. +/=/- | CEC-2017 test suite | | | | CEC-2022 test suite | |
|---|---|---|---|---|---|---|
| | 10D | 30D | 50D | 100D | 10D | 20D |
| ECO | 24/2/3 | 28/1/0 | 24/3/2 | 22/3/4 | 10/1/1 | 11/1/0 |
| SAO | 17/6/6 | 11/9/9 | 16/2/11 | 22/5/2 | 7/4/1 | 4/3/5 |
| CFOA | 22/4/3 | 27/1/1 | 27/1/1 | 29/0/0 | 9/2/1 | 9/3/0 |
| DBO | 23/4/2 | 28/1/0 | 27/2/0 | 27/2/0 | 8/3/1 | 9/1/2 |
| QIO | 17/4/8 | 21/6/2 | 19/4/6 | 16/4/9 | 8/3/1 | 9/2/1 |
| AE | 21/5/3 | 19/3/7 | 15/2/12 | 11/5/13 | 9/2/1 | 7/0/5 |

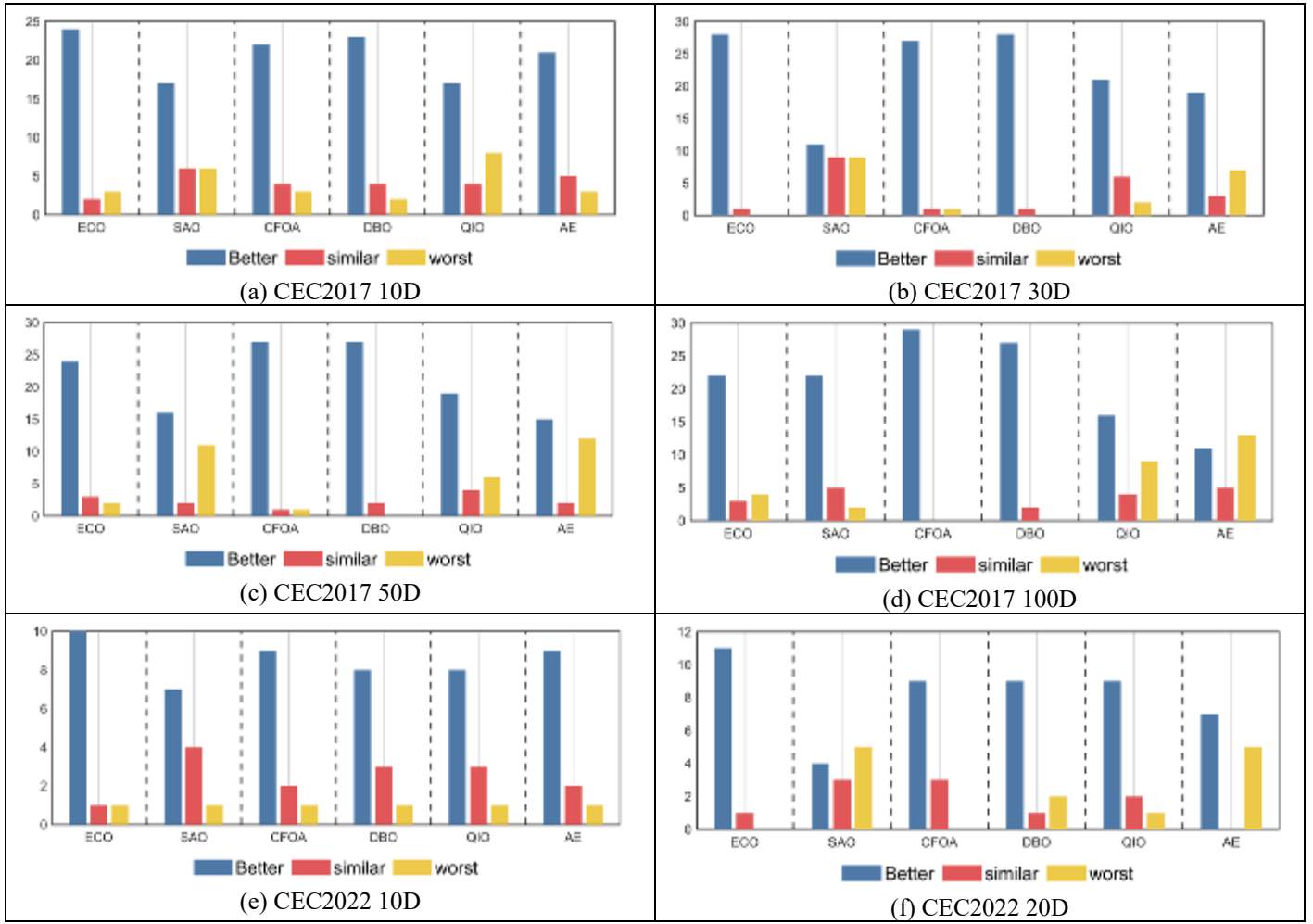

**Figure 7.** The visualization of the Wilcoxon rank sum test results of IECO-MCO and other basic competitors

4.3.3. Analysis of the Friedman test results

Table 7 presents the Friedman test results for IECO-MCO and the basic competitors on the CEC2017 and CEC2022 test sets. The ranking trends of each algorithm are illustrated in Figure 8. According to Table 7, the p-values for all scenarios are less than 0.05, indicating significant differences between the basic algorithms and IECO-MCO involved in the experiments. IECO-MCO achieves the best Friedman rankings of 2.241, 2.103, 2.276, 2.241, 2.000, and 2.417 across all scenarios, compared to the basic ECO's rankings of 4.966, 5.000, 4.655, 3.897, 5.417, and 5.750, highlighting a substantial performance gap between IECO-MCO and the basic RIME. Furthermore, as shown in Figure 8, the Friedman rankings of IECO-MCO exhibit minimal fluctuations across the four scenarios, demonstrating its insensitivity to dimensional changes and thus greater scalability. Additionally, based on the earlier Wilcoxon rank-sum test results, although AE and SAO outperform IECO-MCO in certain dimensions with more functions where they excel than those where they underperform, their average Friedman rankings in those dimensions are lower than those of IECO-MCO. This suggests that SAO and AE have limited scalability and excel only in specific problems, whereas IECO-MCO demonstrates superior adaptability and holds broader application potential.

**Table 7.** The Friedman test results of IECO-MCO and other basic competitors

| Algorithm | CEC-2017 test suite | | | | | CEC-2022 test suite | | | Overall average ranking |
|---|---|---|---|---|---|---|---|---|---|
| | 10D | 30D | 50D | 100D | Average ranking | 10D | 20D | Average ranking | |
| IECO-MCO | **2.241** | **2.103** | **2.276** | **2.241** | **2.216** | **2.000** | **2.417** | **2.208** | **2.213** |
| ECO | 4.966 | 5.000 | 4.655 | 3.897 | 4.629 | 5.417 | 5.750 | 5.583 | 4.947 |
| SAO | 3.759 | 2.552 | 3.655 | 4.931 | 3.724 | 2.833 | 3.000 | 2.917 | 3.455 |
| CFOA | 4.241 | 5.310 | 5.172 | 5.241 | 4.991 | 4.417 | 4.417 | 4.417 | 4.800 |
| DBO | 6.172 | 6.517 | 6.414 | 6.414 | 6.379 | 5.750 | 5.917 | 5.833 | 6.197 |
| QIO | 2.897 | 3.655 | 3.310 | 2.966 | 3.207 | 3.333 | 3.917 | 3.625 | 3.346 |
| AE | 3.724 | 2.862 | 2.517 | 2.310 | 2.853 | 4.250 | 2.583 | 3.417 | 3.041 |
| P-value | 1.05E-11 | 1.05E-11 | 7.62E-16 | 2.97E-18 | N/A | 7.33E-05 | 7.33E-05 | N/A | N/A |

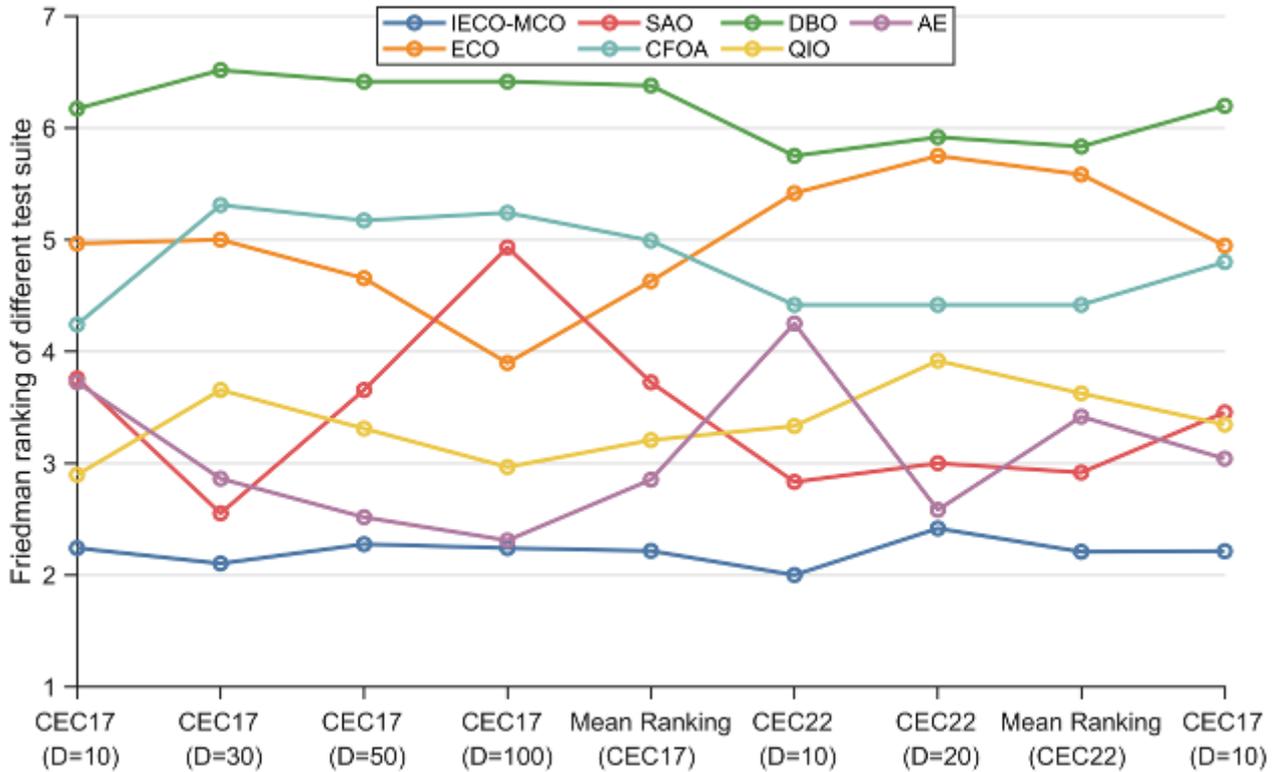

**Figure 8.** The visualization of the Friedman test results of IECO-MCO and other basic competitors

4.3.4. Analysis of convergence behavior

This subsection aims to assess the convergence behavior and speed of IECO-MCO in comparison to other basic algorithms. Figure 9 displays the convergence curves for various function categories, with each curve representing the average of the best outcomes from 30 independent runs of IECO-MCO and its baseline counterparts. Convergence analysis plays a pivotal role in metaheuristic algorithms, as it evaluates the efficacy and efficiency of the optimization process. By examining convergence, one can determine the rate at which an algorithm identifies optimal or near-optimal solutions and the stability of the optimization trajectory, thereby gauging whether the algorithm is advancing toward high-quality solutions. During optimization, IECO-MCO typically manifests three distinct traits. Initially, it undergoes a phase of rapid convergence, where solutions of reasonable quality—though not necessarily optimal—are identified early in the iterations, followed by a plateau with minimal changes over subsequent iterations. Subsequently, IECO-MCO exhibits accelerated convergence in the initial half of the iterations, where substantial improvements to the solutions are realized. Finally, IECO-MCO demonstrates steady improvements in later iterations, with minor refinements to the solutions until the stopping criteria are met.

Collectively, these traits enable IECO-MCO to achieve a balance between exploration and exploitation throughout the optimization process. By employing three covariance learning operators, the IECO-MCO algorithm achieves comprehensive exploitation and exploration while maintaining adequate population diversity. The findings in Figure 9 reveal that IECO-MCO surpasses other comparative algorithms in terms of rapid convergence and preserving diversity for multimodal and composition functions. As a result, IECO-MCO stands out as a robust candidate for tackling diverse optimization challenges. The exceptional convergence performance of IECO-MCO is attributed to its three unique covariance learning operators. The Gaussian covariance operator directs ECO toward favorable directions, thereby improving the overall population quality. The shift covariance operator adjusts the movement direction of elite agents using multiple reference points, ensuring varied movement directions for each agent and achieving an optimal balance between exploration and exploitation. The differential covariance operator sustains population diversity by utilizing variations between random solutions throughout the optimization process.

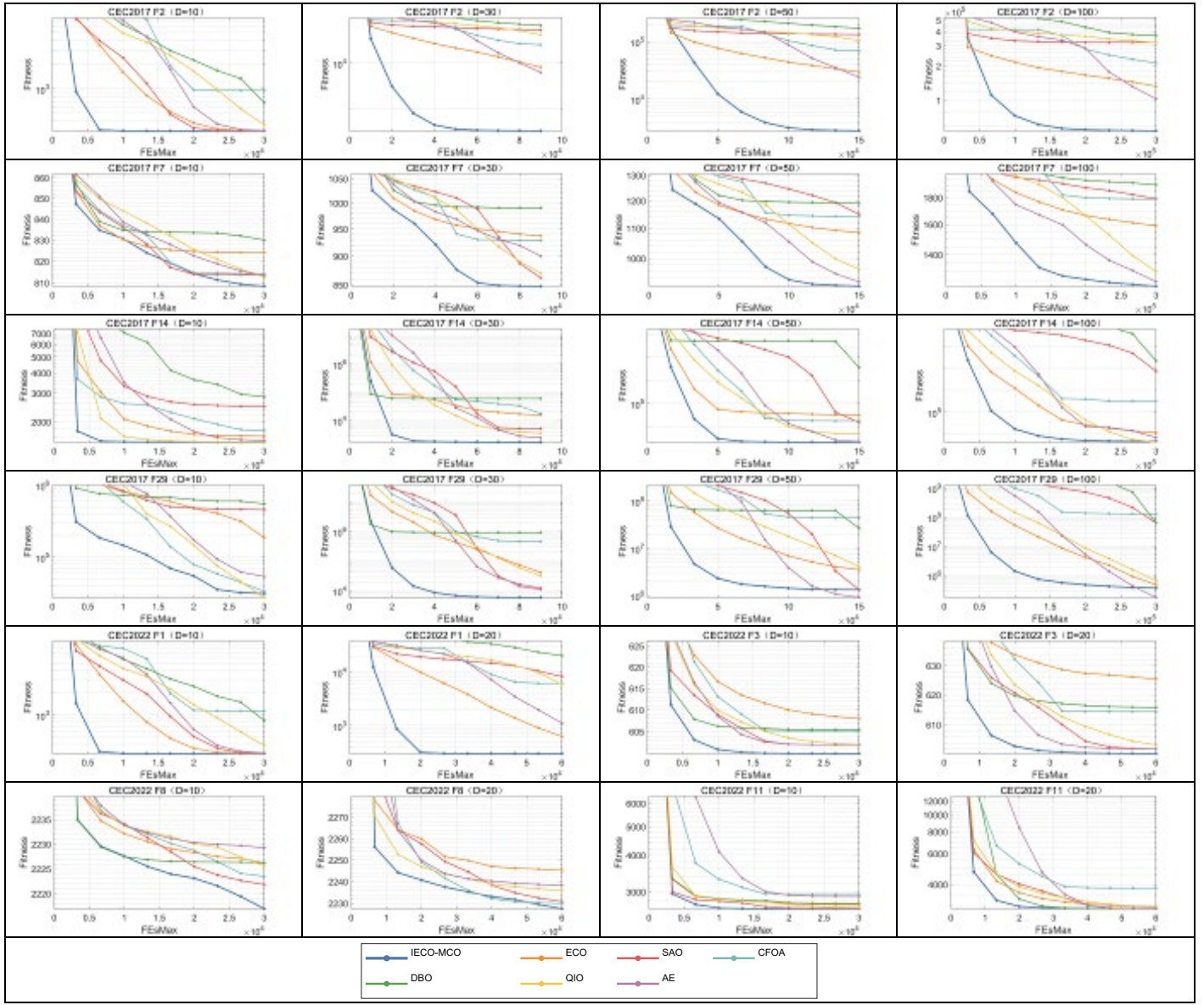

**Figure 9.** Convergence curves of IECO-MCO and other basic competitors

*4.4. Comparison with other improved algorithms using CEC2017/CEC2022*

In Section 4.3, this paper compares IECO-MCO with advanced basic algorithms, confirming its superior performance. In this subsection, to further evaluate the capabilities of IECO-MCO, five variants of metaheuristic algorithms from different categories are selected for comparison with the proposed IECO-MCO. These variants include RDGMVO, ISGTOA, AFDBARO, MTVSCA, and ALSHADE. To eliminate dispute and ensure a fair comparison, the parameter configurations for the competitors follow the values suggested in their source literature, as outlined in Table 2. The detailed analysis results are presented below.

4.4.1. Analysis of quantitative results

The full results, including best values, standard deviation, and mean values obtained by IECO-MCO and the improved competitor on the CEC2017 and CEC2022 test sets, are displayed in Table S7-Table S12 in Table S1-Table S6 in supplementary files. In this subsection, the rankings of each algorithm based on mean values for each function are briefly illustrated using radar charts, as shown in Figure 10. In the radar charts, each algorithm forms a polygon based on its rankings. The smaller the area of the polygon, the better the overall performance of the algorithm. To better observe the areas of IECO-MCO and ECO, the rankings of the top three algorithms—IECO-MCO, ISGTOA, and ALSHADE—along with the basic ECO, are displayed. According to Figure 7, it is evident that the area of the proposed IECO-MCO is relatively smaller, indicating that IECO-MCO exhibits the best performance. Subsequently, statistical tests will be employed to analyze the experimental data obtained by IECO-MCO and the improved competitors, ensuring accurate conclusions and avoiding potential inaccuracies arising from statistical averages of the results.

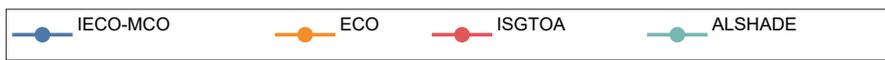

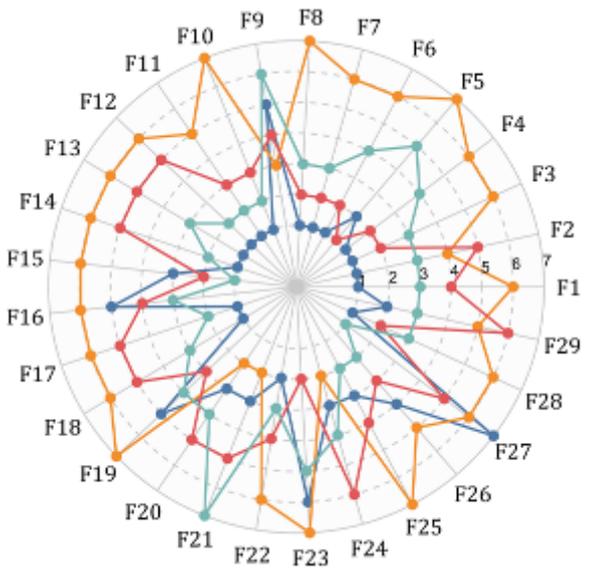

(a) CEC2017 10D

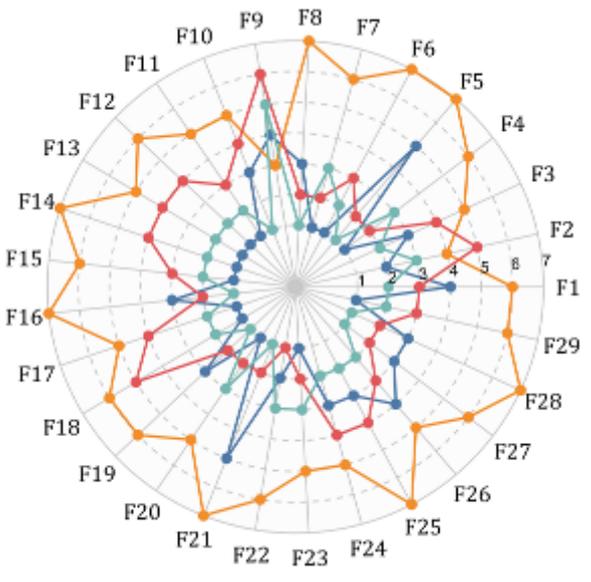

(b) CEC2017 30D

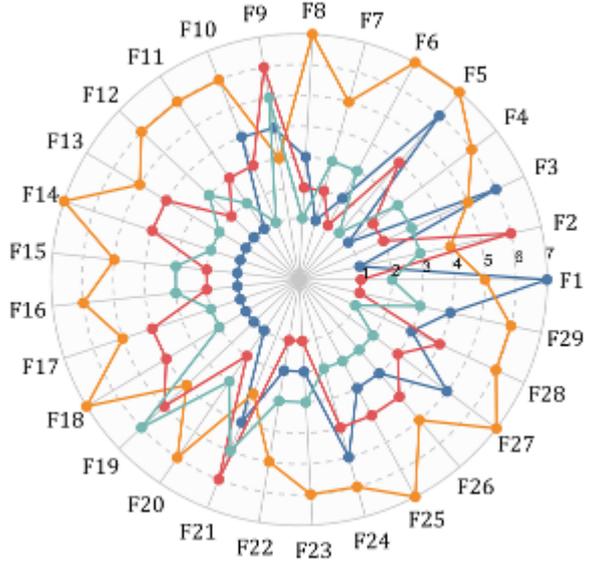

(c) CEC2017 50D

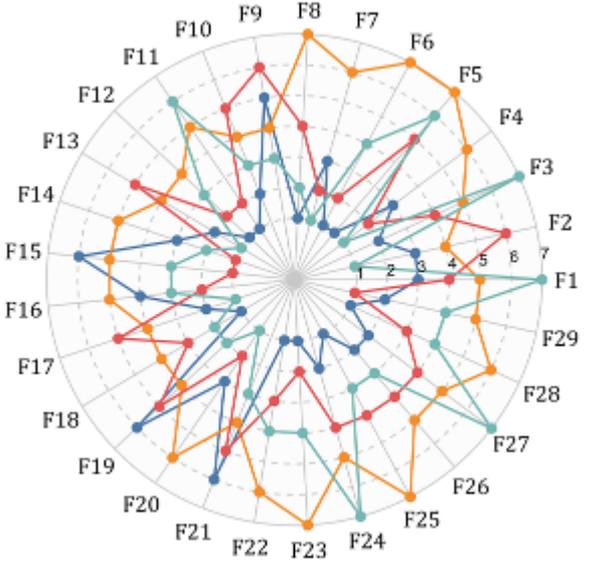

(d) CEC2017 100D

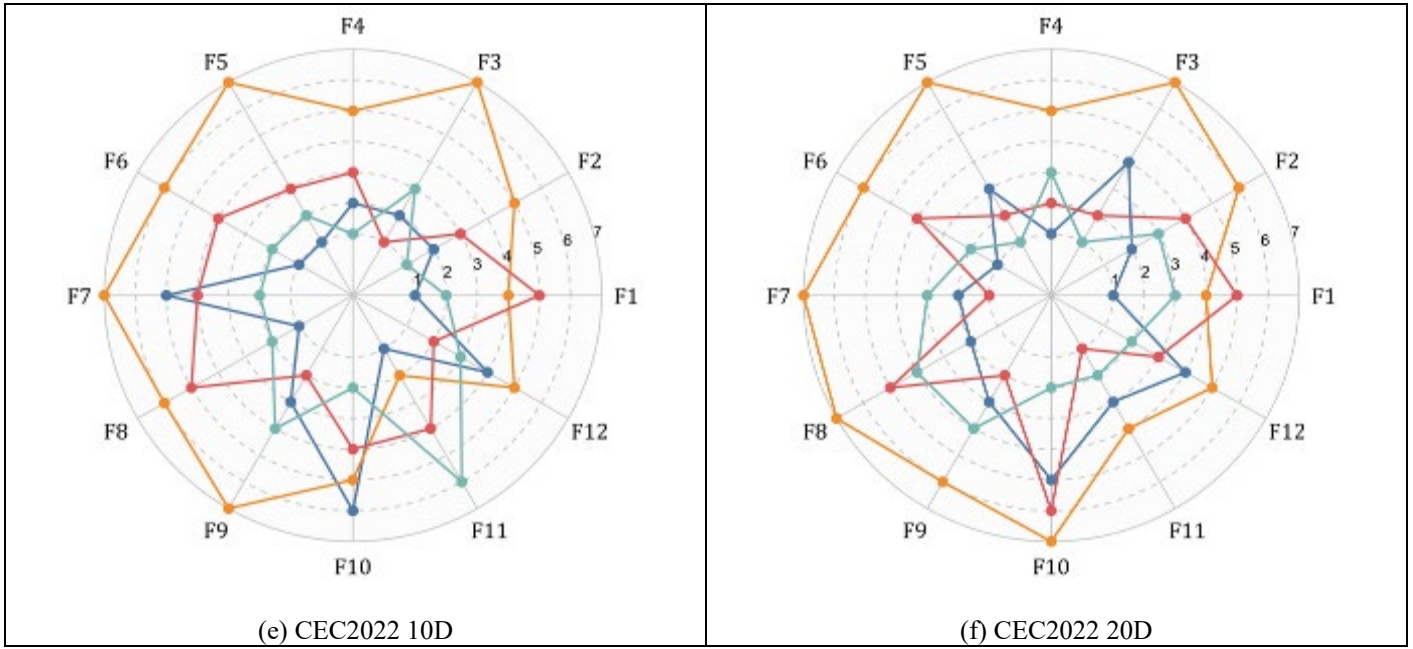

| (e) CEC2022 10D | (f) CEC2022 20D |

**Figure 10.** Rankings of IECO-MCO and other improved algorithms based on mean values on each benchmark function

4.4.2. Analysis of the Kruskal-Wallis test results

The Kruskal-Wallis test is utilized to analyze the experimental results of IECO-MCO and its competitors on the CEC2017 and CEC2022 test sets. The Kruskal-Wallis test results across different dimensions are depicted in Figure 11. IECO-MCO yields the second-lowest rank values in CEC2017 30D and 50D, while achieving the lowest rank values in all other scenarios. Notably, the average ranks of MRIMR-CD exhibit minimal fluctuations across varying dimensions, indicating robust adaptability. Furthermore, Table S13– Table S18 in Table S1-Table S6 in the supplementary files provide the p-values from the Kruskal-Wallis tests. Similar to the Wilcoxon rank-sum test, when the p-value exceeds 0.05, no significant difference exists between IECO-MCO and the comparison algorithms. Conversely, when the p-value is less than 0.05, a significant difference is observed between IECO-MCO and the comparison algorithms. Based on the Kruskal-Wallis test p-values, significant differences are evident between IECO-MCO and its competitors. These results are illustrated in Figure 12.

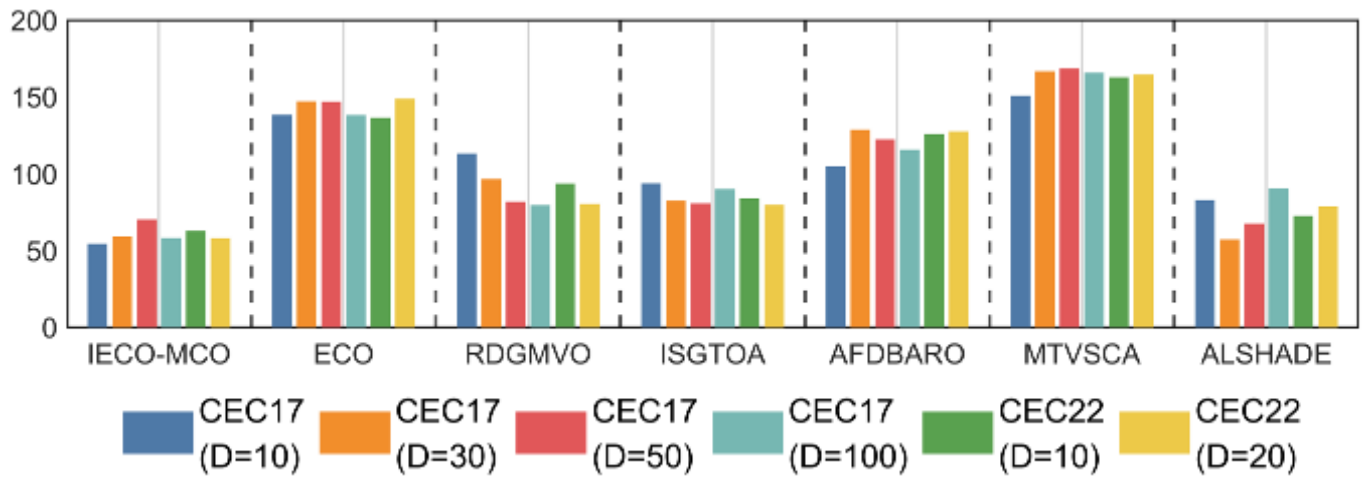

**Figure 11.** Kruskal-Wallis ranking of IECO-MCO and other improved algorithms

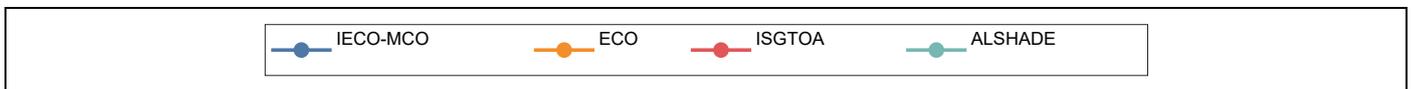

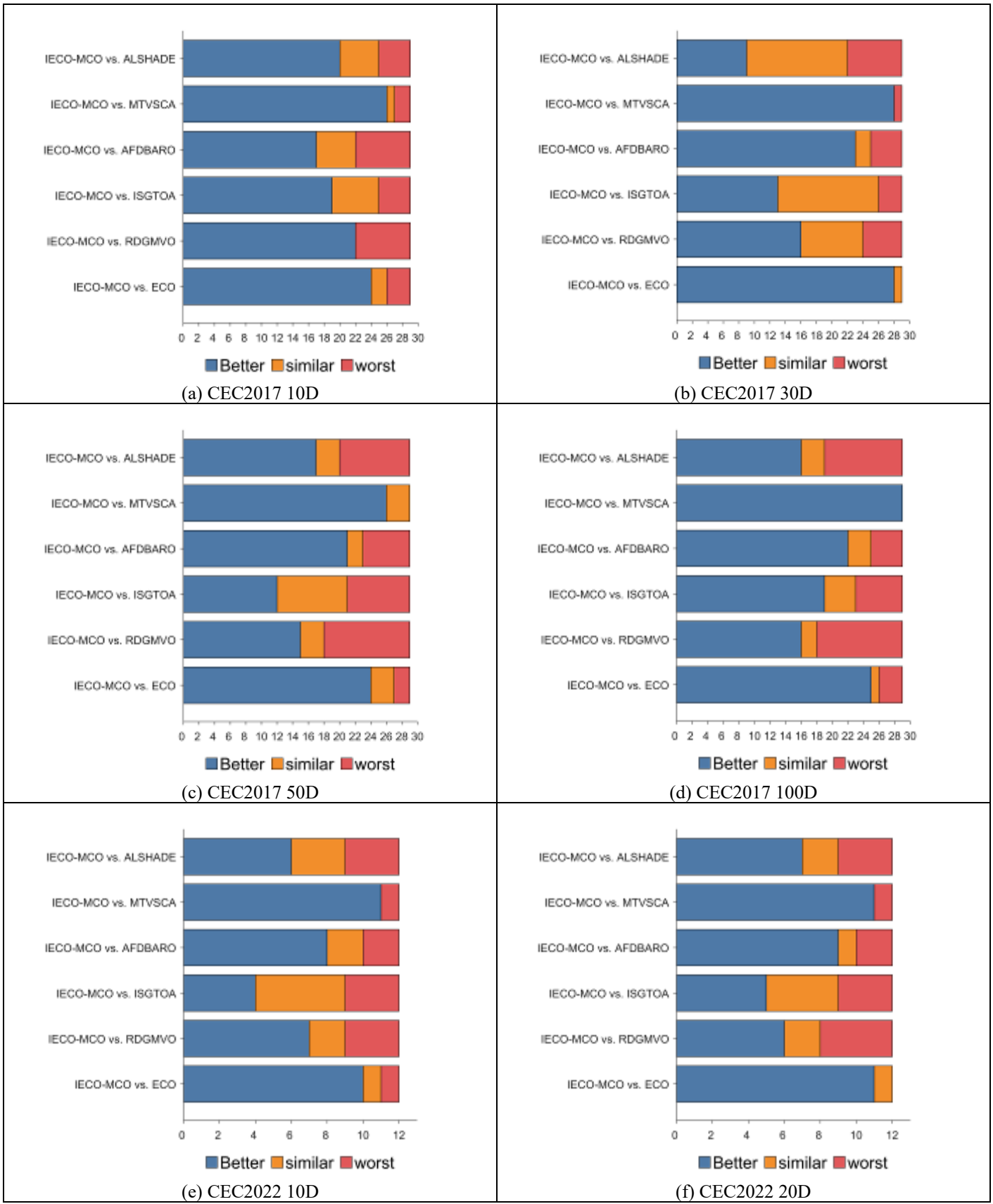

**Figure 12.** The visualization of the Kruskal-Wallis test results of IECO-MCO and other improved algorithms

4.4.3. Analysis of the Friedman test results

In this section, the results of the Friedman test are analyzed to determine significant differences among multiple algorithmic approaches. Table 8 provides the average rankings of IECO-MCO and other competing algorithms across six scenarios. Figure 13 further supports this analysis, offering a visual representation of the rankings based on the Friedman test, clearly demonstrating that IECO-MCO outperforms other improved algorithms in most cases. According to the test results, the proposed IECO-MCO achieves the best performance in CEC2017 10D and 100D, as well as in CEC2022 10D and 20D. It ties for first place with AL-SHADE in CEC2017 50D and ranks second in CEC2017 30D. The proposed IECO-MCO secures the top position across both test sets, achieving average rankings of 2.422 and 2.500, while ALSHADE ranks second in the CEC2017 test set and ties for first place with IECO-MCO in the CEC2022 test set. The details of the Friedman test results are as follows.

For CEC2017 10D, IECO-MCO ranks in the first place, followed by ALSHADE, AFDBARO, ISGTOA, RDGMVO, MTVSCA, and ECO. That is, IECO-MCO outperforms five improved competitors on CEC2017 10D functions.

For CEC2017 30D, ALSHADE ranks in the first place, followed by IECO-MCO, ISGTOA, RDGMVO, AFDBARO, ECO, and MTVSCA. That is, IECO-MCO outperforms five improved competitors except ALSHADE on CEC2017 30D functions.

For CEC2017 50D, IECO-MCO and ALSHADE share the first place, followed by ISGTOA, RDGMVO, AFDBARO, ECO, and MTVSCA. That is, IECO-MCO outperforms five improved competitors except ALSHADE on CEC2017 50D functions.

For CEC2017 100D, IECO-MCO ranks in the first place, followed by RDGMVO, ALSHADE/ISGTOA, AFDBARO, ECO, and MTVSCA. That is, IECO-MCO outperforms five improved competitors on CEC2017 100D functions.

For CEC2022 10D, IECO-MCO ranks in the first place, followed by ALSHADE, ISGTOA, RDGMVO, AFDBARO, ECO, and MTVSCA. That is, IECO-MCO outperforms five improved competitors on CEC2022 10D functions.

For CEC2022 20D, ALSHADE ranks in the first place, followed by IECO-MCO, ISGTOA, RDGMVO, AFDBARO, MTVSCA, and ECO. That is, IECO-MCO outperforms five improved competitors except ALSHADE on CEC2022 20D functions.

Based on the above discussions, IECO-MCO is superior to ISGTOA, RDGMVO, AFDBARO, MTVSCA, and ECO in all cases, superior to ALSHADE on CEC2017 10D/100D and CEC2022 10D, but is inferior to it on CEC2017 30D and CEC2022 20D. According to the "Overall average ranking" in Table 8, it can be inferred that IECO-MCO surpasses all competing algorithms, indicating that IECO-MCO substantially enhances the efficacy of the ECO algorithm, and IECO-MCO is a promising variant of ECO.

Table 8. The Friedman test results of IECO-MCO and other improved competitors

| Algorithm | CEC-2017 test suite | | | | | CEC-2022 test suite | | | Overall average ranking |
|---|---|---|---|---|---|---|---|---|---|
| | 10D | 30D | 50D | 100D | Average ranking | 10D | 20D | Average ranking | |
| IECO-MCO | **2.345** | **2.276** | **2.621** | **2.448** | **2.422** | **2.417** | **2.583** | **2.500** | **2.448** |
| ECO | 5.517 | 5.793 | 5.621 | 5.207 | 5.534 | 5.500 | 5.917 | 5.708 | 5.592 |
| RDGMVO | 4.655 | 4.138 | 3.276 | 3.069 | 3.784 | 4.250 | 3.333 | 3.792 | 3.787 |
| ISGTOA | 3.655 | 3.069 | 3.069 | 3.379 | 3.293 | 3.333 | 3.083 | 3.208 | 3.265 |
| AFDBARO | 3.276 | 4.655 | 4.655 | 4.552 | 4.284 | 4.333 | 4.750 | 4.542 | 4.370 |
| MTVSCA | 5.379 | 6.034 | 6.138 | 5.966 | 5.879 | 5.667 | 5.833 | 5.750 | 5.836 |
| ALSHADE | 3.172 | 2.034 | 2.621 | 3.379 | 2.802 | 2.500 | 2.500 | 2.500 | 2.701 |
| P-value | 7.13E-10 | 1.44E-18 | 4.15E-15 | 4.32E-11 | N/A | 1.61E-04 | 7.40E-06 | N/A | N/A |

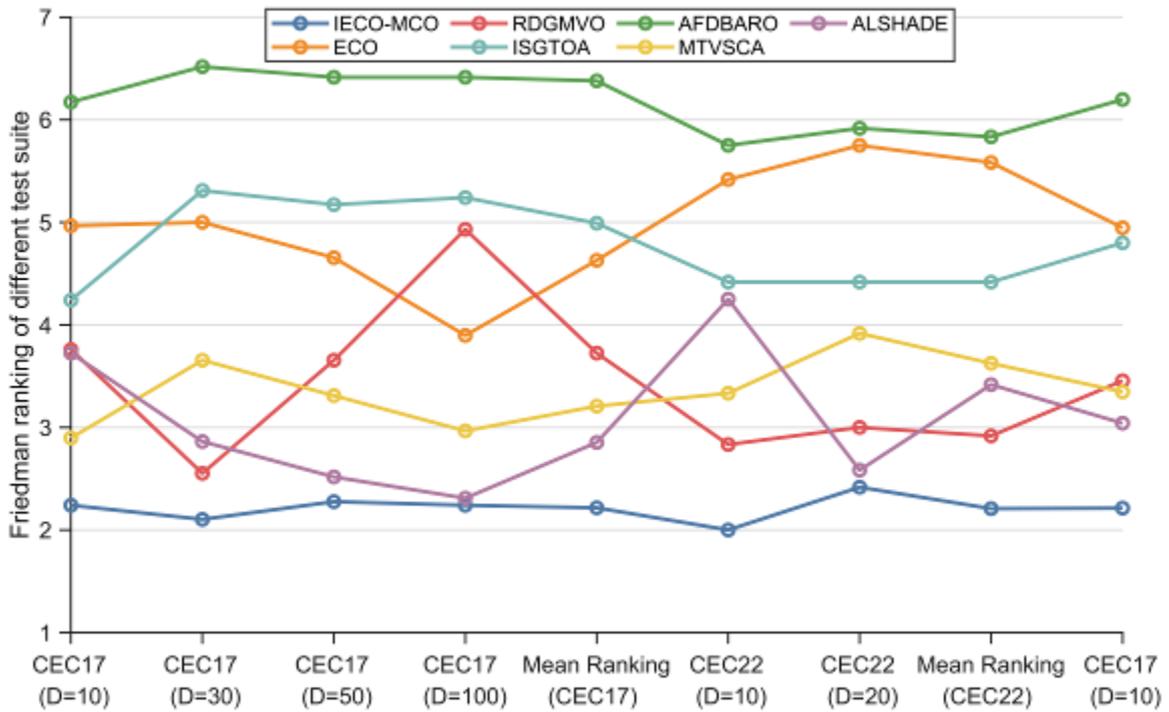

**Figure 13.** The visualization of the Friedman test results of IECO-MCO and other improved competitors

4.4.4. Analysis of robustness behavior

    This subsection aims to evaluate the robustness of IECO-MCO in comparison to other enhanced algorithms. Robustness analysis is of significant importance in metaheuristic algorithms, as it facilitates the assessment of optimization process stability. By conducting robustness analysis, it is possible to determine whether an algorithm exhibits better potential for practical applications, given that real-world scenarios prioritize stability and sustainability. Figure 14 presents box plots for various function categories. The central line in each boxplot represents the median of the experimental results, reflecting the average level of the sample data. The upper and lower boundaries of the boxplot correspond to the upper and lower quartiles of the experimental data, respectively. The "+" symbol indicates outliers. In the boxplot, a narrower box signifies more stable experimental data. As shown in the figure, the boxplot generated from IECO-MCO results contains fewer outliers and narrower boxes, demonstrating that the solutions provided by IECO-MCO are more concentrated and its problem-solving capability is more stable compared to other competing algorithms. This indicates that the three covariance operators proposed in this study enhance the performance of IECO-MCO without compromising its robustness.

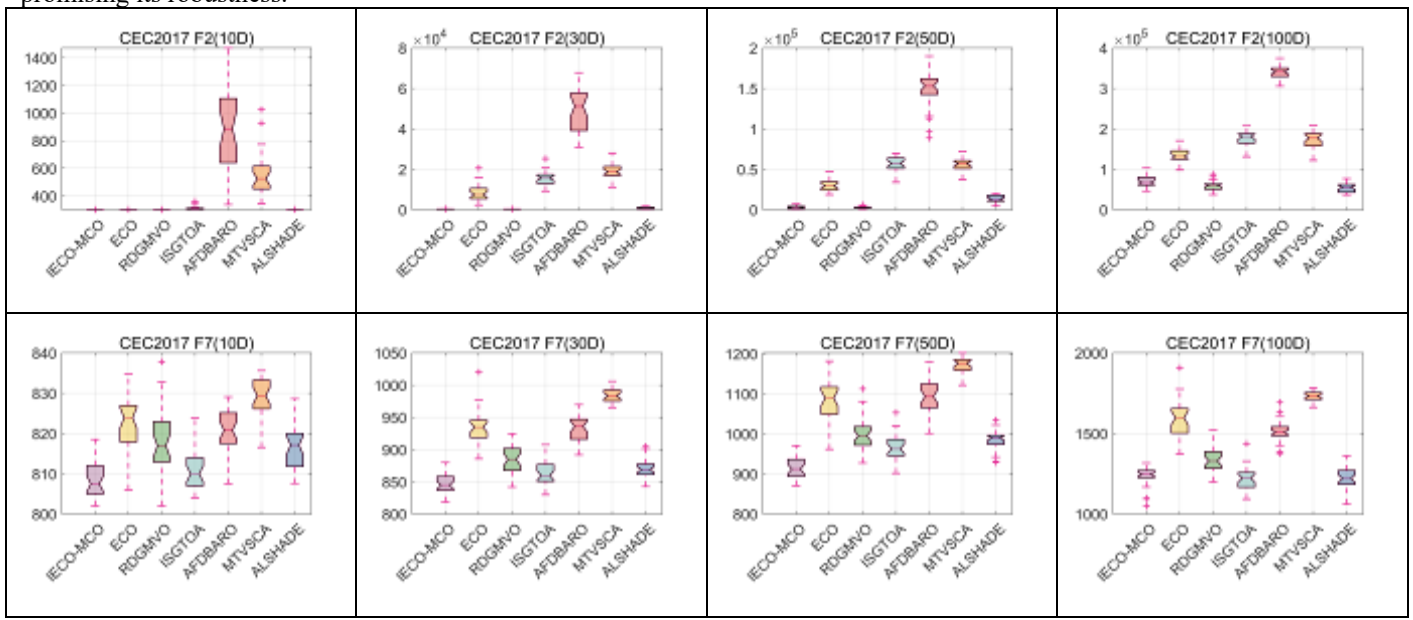

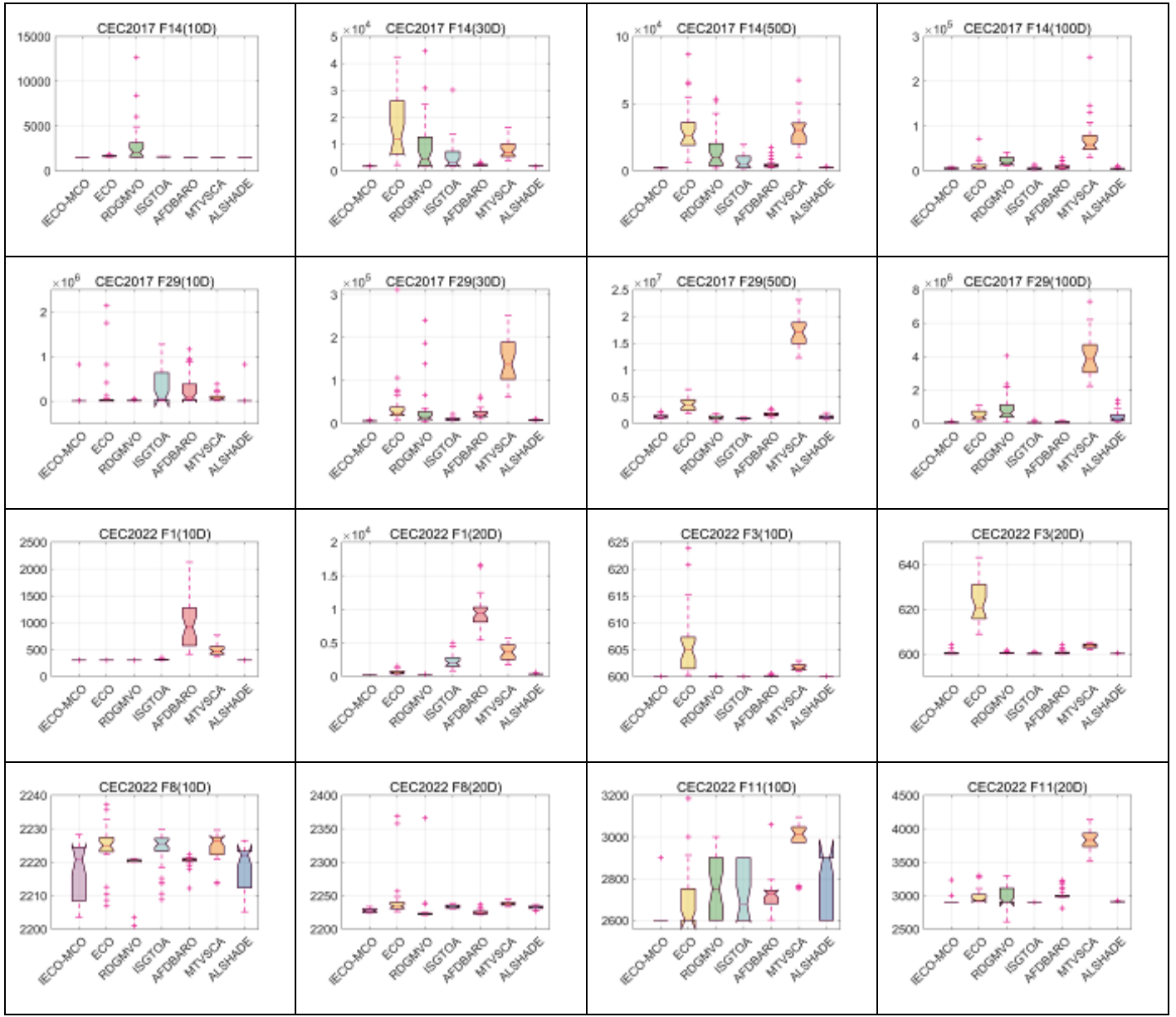

**Figure 14.** Box graph of IECO-MCO and other improved competitors

## 5. Engineering design optimization problems

In this subsection, the effectiveness of the IECO-MCO algorithm is evaluated through experiments on ten engineering design optimization problems. The top three basic algorithms (AE and QIO) and improved algorithms (ISGTOA and ALSHADE) from previous experiments, along with the basic ECO algorithm, are selected as competitors. Table 9 provides a summary of these engineering design optimization problems. Due to the numerous constraints and the complexity of the objective functions, real-world engineering optimization problems are inherently challenging. Over time, various constraint-handling techniques have been developed to address the limitations associated with penalty functions. This study employs the penalty function method, incorporating a conditional statement after each iteration to assess constraint satisfaction. If constraints are not met, the population individuals are reinitialized. The experimental and algorithmic parameter settings remain consistent with those described earlier. The best values, mean values, and standard deviations obtained from 30 independent runs are presented in Table 10.

**Table 9.** Ten real-world constrained engineering optimization problems

| Problem | Name | D |
|---|---|---|
| RW01 | Tension/compression spring design problem | 3 |
| RW02 | Pressure vessel design problem | 4 |
| RW03 | Three-bar truss design problem | 2 |

|       | RW04 | Welded beam design problem | 4 |
|-------|------|---------------------------|---|
|       | RW05 | Speed reducer design problem | 7 |
|       | RW06 | Gear train design problem | 4 |
|       | RW07 | Rolling element bearing design | 10 |
|       | RW08 | Cantilever beam design problem | 5 |
|       | RW09 | Multiple disk clutch brake design problem | 5 |
|       | RW10 | Step-cone pulley problem | 5 |

**Table 10.** Results of ten real-world constrained optimization problems obtained by IECO-MCO and other competitors

| Problem ID | Index | IECO-MCO | ECO | QIO | AE | ISGTOA | ALSHADE |
|---|---|---|---|---|---|---|---|
| RW1 | Best | 1.2667E-02 | 1.2672E-02 | 1.2680E-02 | 1.2683E-02 | 1.2686E-02 | 1.2666E-02 |
|     | Mean | 1.2712E-02 | 1.3142E-02 | 1.2732E-02 | 1.2764E-02 | 1.2732E-02 | **1.2699E-02** |
|     | Std  | 7.4463E-05 | 7.2167E-04 | 6.0922E-05 | 5.2830E-05 | 4.9854E-05 | 4.2592E-05 |
|     | Rank | 2 | 6 | 4 | 5 | 3 | 1 |
| RW2 | Best | 5.8701E+03 | 5.9436E+03 | 5.8991E+03 | 5.8991E+03 | 5.8701E+03 | 5.8707E+03 |
|     | Mean | **5.8763E+03** | 6.6303E+03 | 5.9875E+03 | 5.9666E+03 | 5.8991E+03 | 5.9051E+03 |
|     | Std  | 3.3850E+01 | 4.8333E+02 | 8.3981E+01 | 5.3198E+01 | 4.5167E+01 | 4.9379E+01 |
|     | Rank | 1 | 6 | 5 | 4 | 2 | 3 |
| RW3 | Best | 2.6389E+02 | 2.6389E+02 | 2.6389E+02 | 2.6389E+02 | 2.6389E+02 | 2.6389E+02 |
|     | Mean | 2.6389E+02 | 2.6394E+02 | **2.6389E+02** | 2.6389E+02 | 2.6389E+02 | 2.6389E+02 |
|     | Std  | 5.9574E-08 | 1.3885E-01 | 1.7843E-08 | 1.8459E-05 | 7.1785E-06 | 1.3934E-06 |
|     | Rank | 2 | 6 | 1 | 5 | 4 | 3 |
| RW4 | Best | 1.6928E+00 | 1.7308E+00 | 1.6950E+00 | 1.6932E+00 | 1.6928E+00 | 1.6930E+00 |
|     | Mean | 1.6932E+00 | 1.9560E+00 | 1.6987E+00 | 1.6935E+00 | **1.6929E+00** | 1.6956E+00 |
|     | Std  | 1.9926E-04 | 1.9693E-01 | 2.7950E-03 | 1.9780E-04 | 1.3918E-04 | 3.3571E-03 |
|     | Rank | 2 | 6 | 5 | 3 | 1 | 4 |
| RW5 | Best | 2.9936E+03 | 2.9936E+03 | 2.9953E+03 | 2.9937E+03 | 2.9936E+03 | 2.9936E+03 |
|     | Mean | 2.9936E+03 | 2.9997E+03 | 2.9975E+03 | 2.9937E+03 | **2.9936E+03** | 2.9936E+03 |
|     | Std  | 9.6021E-06 | 5.2885E+00 | 1.0024E+00 | 3.6051E-02 | 4.8510E-13 | 2.2492E-03 |
|     | Rank | 2 | 6 | 5 | 4 | 1 | 3 |
| RW6 | Best | 2.7009E-12 | 2.7009E-12 | 2.7009E-12 | 2.7009E-12 | 2.7009E-12 | 2.7009E-12 |
|     | Mean | 8.1547E-11 | 7.7978E-10 | 9.7236E-11 | **2.2070E-11** | 2.6605E-10 | 5.8297E-10 |
|     | Std  | 2.3622E-10 | 7.3166E-10 | 2.5834E-10 | 3.7106E-11 | 4.1674E-10 | 8.8552E-10 |
|     | Rank | 2 | 6 | 3 | 1 | 4 | 5 |
| RW7 | Best | -2.4358E+05 | -2.4358E+05 | -2.4340E+05 | -2.5047E+05 | -2.4358E+05 | -2.4358E+05 |
|     | Mean | -2.4358E+05 | -2.4358E+05 | -2.4250E+05 | -2.6906E+04 | **-2.4358E+05** | -2.4358E+05 |
|     | Std  | 7.5275E-11 | 9.1397E-11 | 6.7549E+02 | 5.9648E+04 | 8.9948E-11 | 9.5504E-06 |
|     | Rank | 2 | 2 | 5 | 6 | 1 | 4 |
| RW8 | Best | 1.3400E+00 | 1.3406E+00 | 1.3400E+00 | 1.3400E+00 | 1.3400E+00 | 1.3400E+00 |
|     | Mean | **1.3400E+00** | 1.3502E+00 | 1.3403E+00 | 1.3400E+00 | 1.3400E+00 | 1.3404E+00 |
|     | Std  | 6.9961E-07 | 7.4982E-03 | 4.3688E-04 | 2.2233E-05 | 2.3810E-05 | 6.9457E-04 |
|     | Rank | 1 | 6 | 4 | 3 | 2 | 5 |
| RW9 | Best | 3.9247E+12 | 3.9247E+12 | 3.9247E+12 | 2.4780E+12 | 3.9247E+12 | 3.9247E+12 |
|     | Mean | 3.9247E+12 | 3.9247E+12 | 3.9247E+12 | **2.5144E+12** | 3.9247E+12 | 3.9247E+12 |
|     | Std  | 2.9798E-03 | 2.9798E-03 | 2.0334E-02 | 9.9674E+09 | 2.9798E-03 | 2.9798E-03 |
|     | Rank | 2 | 2 | 6 | 1 | 2 | 2 |
| RW10 | Best | 1.6086E+01 | 1.6091E+01 | 1.6093E+01 | 1.6105E+01 | 1.6091E+01 | 1.6086E+01 |
|      | Mean | **1.6086E+01** | 1.6653E+01 | 1.6310E+01 | 1.6155E+01 | 1.6196E+01 | 1.6107E+01 |
|      | Std  | 5.0346E-06 | 3.5327E-01 | 1.8503E-01 | 6.2808E-02 | 1.0553E-01 | 8.1386E-02 |
|      | Rank | 1 | 6 | 5 | 3 | 4 | 2 |

Table 10 demonstrates the effectiveness of the IECO-MCO algorithm in solving most engineering design optimization problems, showcasing robust performance. Notably, on the 10 real-world engineering optimization problems considered, the proposed IECO-MCO algorithm consistently ranks within the top two. Specifically, IECO excels in solving problems RW02, RW08, and RW10. QIO provides the best results for RW03. AE ranks first on RW06 and RW09, while ISGTOA delivers satisfactory solutions for RW04, RW05, and RW07. ALSHADE performs best on RW01. IECO-MCO ranks second on these seven problems, indicating its potential to further address real-world optimization challenges. Table 11 summarizes the results of three statistical tests, with the p-values from the Friedman test and Kruskal-Wallis test both being less than 0.05, indicating significant differences between IECO-MCO and its competitors. Specifically, IECO achieves a mean Friedman rank of 1.90, compared to 4.30 for QIO, 3.50 for AE$\epsilon$, 2.55 for ISGTOA, 3.35 for ALSHADE, and 5.40 for the basic ECO. Thus, IECO-MCO exhibits superior performance relative to the other four leading algorithms. Additionally, IECO-MCO achieves the lowest rank value with 47.42, while the basic ECO algorithm yields the highest value with 135.19. In conclusion, IECO-MCO successfully addresses a variety of real-world problems, demonstrating satisfactory performance. Its overall performance is highly competitive compared to other algorithms.

**Table 11.** Statistical results of IECO-MCO and other competitors

| Algorithm | IECO-MCO | ECO | QIO | AE | ISGTOA | ALSHADE | P-value |
|---|---|---|---|---|---|---|---|
| Friedman test | **1.90** | 5.40 | 4.30 | 3.50 | 2.55 | 3.35 | 1.90 |
| Wilcoxon rank-sum test | IECO-MCO vs. +/=/- | 8/1/1 | 7/3/0 | 8/1/1 | 6/1/3 | 8/2/0 | N/A |
| Kruskal Wallis test | **47.42** | 135.19 | 109.13 | 94.02 | 71.83 | 85.41 | N/A |

## 6. Conclusions

This study introduces IECO-ECO, a novel variant of ECO. The proposed IECO-ECO integrates three covariance learning operators: the Gaussian covariance operator, the shift covariance operator, and differential covariance operator, each contributing to the overall enhancement of the algorithm's performance.

**Gaussian Covariance Operator:** This operator guides the population toward advantageous directions, thereby improving the overall population quality. By promoting broader exploration, it enhances search efficiency, enabling the algorithm to better understand the global landscape of the search space.

**Shift Covariance Operator:** This operator adjusts the movement direction of elite agents using multiple reference points, ensuring that each agent follows a distinct trajectory. This approach achieves an appropriate balance between exploration and exploitation.

**Differential Covariance Operator:** By leveraging the differences between random solutions, this operator maintains population diversity during the optimization process, thereby enhancing the robustness of the ECO framework.

The effectiveness of IECO-MCO has been evaluated across multiple domains, including 42 test functions from the CEC 2017 and CEC 2022 test suites, and 10 engineering design optimization problems. IECO-MCO demonstrates exceptional performance, surpassing its own derived variants (GECO, SECO, DECO), five basic algorithms from different categories (SAO, CFOA, AE, DBO, QIO), and improved algorithms from various categories (RDGMVO, ISGTOA, AFDBARO, MTVSCA, ALSHADE). These results underscore the effectiveness and superiority of IECO-MCO in addressing a wide range of optimization challenges. Advances in convergence speed, search capability, and robustness position the IECO-MCO algorithm as a highly competitive contender in the field of optimization.

Although IECO-MCO demonstrates promising optimization capabilities, as optimization problems continue to grow in complexity, there are still aspects of its design principles and methodologies that can be further enhanced. (1) Application to complex optimization problems: Future work could extend its application to areas such as mission planning (trajectory planning and target allocation), medical image segmentation, automated artistic design, and large-scale model architecture optimization. We will explore methods of constraint handling in solving these problems. (2) Development of additional versions: Real-world problems are diverse and complex, necessitating the development of multi-objective versions of IECO-MCO. Additionally, a binary version represents another promising research direction. The development of multi-objective algorithms helps to efficiently balance multiple conflicting objectives in complex engineering problems and enhance the comprehensive performance and practicality of the solutions. The binary version of the ECO algorithm can efficiently solve discrete problems, improve computational efficiency, enhance global search capability, and adapt to complex engineering scenarios. (3) Further performance enhancement: While the proposed IECO-MCO exhibits excellent overall performance, its suboptimal performance on certain functions motivates further improvements. We plan to refine the existing framework and explore integrating IECO-MCO with other artificial intelligence techniques, such as reinforcement learning and deep learning. For example, we can use reinforcement learning to help IECO-MCO to choose a search strategy to update the location based on the score of improved strategies, rather than based on the optimization process. (4) Generalization of multi-covariance operators: The multi-covariance operators proposed in this study have proven effective in enhancing ECO's performance. However, exploring their applicability to other metaheuristic algorithms may yield deeper insights and advancements.

**Competing Interest**

The authors declare that the authors have no competing interests as defined by Nature Research, or other interests that might be perceived to influence the results and/or discussion reported in this paper.

**Data Availability Statement**

The data is provided within the manuscript.

**Conflicts of Interest**

The authors declare no conflict of interest.

**Author contributions**

**Baoqi Zhao**: conceptualization, methodology, writing, data testing, reviewing, software. **Xiong Yang**: methodology, conceptualization, supervision, formal analysis. **Hoileong Lee:** reviewing, formal analysis. **Bowen Dong**: reviewing, formal analysis.

**Funding**

This research was funded by Ningbo Natural Science Foundation, grant number 2023J242 and Key Project of Ningbo Polytechnic, grant number NZ23Z01.